\documentclass[11pt]{article}

\usepackage{fullpage}
\usepackage[ruled, linesnumbered, vlined, commentsnumbered]{algorithm2e}
\usepackage{graphicx,subfig} % figure related
\usepackage{amsfonts,amssymb,amsmath,amsthm,amsopn}	% math related
\usepackage{hhline,booktabs,colortbl,multirow,tabularx,threeparttable} % table related
\usepackage[listings,skins,breakable]{tcolorbox}

\usepackage{enumerate}
\usepackage{authblk}
\usepackage{footnote}
\usepackage{hyperref}
\usepackage{prettyref}
\usepackage{cite}
\usepackage{setspace}
\usepackage{color}
\usepackage{xcolor}  % Required for custom colors
\usepackage{scrpage2}
\usepackage{geometry}

\usepackage{tikz}
\usepackage{pgfplots}
\usetikzlibrary{positioning,shapes,shadows,arrows,calc}
\tikzstyle{component}=[rectangle, draw=black, rounded corners, fill=blue!40, drop shadow, text centered, anchor=north, text=white, minimum height=1cm]
\tikzstyle{arrow}=[->, thick]

\pgfplotsset{compat=1.12}
\usetikzlibrary{intersections}
\usetikzlibrary{pgfplots.statistics}
\usepgfplotslibrary{fillbetween}

\def\hlinew#1{%
  \noalign{\ifnum0=`}\fi\hrule \@height #1 \futurelet
   \reserved@a\@xhline}
\makeatother

\definecolor{myblue}{RGB}{34,31,217}
\definecolor{mycyan}{gray}{.7}
\definecolor{Gray}{gray}{0.9}

\newcommand{\pref}{\prettyref}

\newrefformat{fig}{Fig.~\ref{#1}}
\newrefformat{tab}{Table~\ref{#1}}
\newrefformat{sec}{Section~\ref{#1}}
\newrefformat{app}{Appendix~\ref{#1}}
\newrefformat{alg}{Algorithm~\ref{#1}}
\newrefformat{property}{Property~\ref{#1}}
\newrefformat{theorem}{Theorem~\ref{#1}}
\newrefformat{corollary}{Corollary~\ref{#1}}
\newrefformat{proposition}{Proposition~\ref{#1}}
\newrefformat{def}{Definition~\ref{#1}}
\newrefformat{eq}{equation~(\ref{#1})}

\title{\vspace{-1ex}\LARGE\textbf{Does Preference Always Help? A Holistic Study on Preference-Based Evolutionary Multi-Objective Optimisation Using Reference Points}\footnote{This manuscript is currently under peer review for possible publication. The reviewer can use this version interchangeably.}}

\author[1]{\normalsize Ke Li}
\author[2]{\normalsize Minhui Liao}
\author[3]{\normalsize Kalyanmoy Deb}
\author[1]{\normalsize Geyong Min}
\author[4,5]{\normalsize Xin Yao}
\affil[1]{\normalsize Department of Computer Science, University of Exeter, EX4 4QF, Exeter, UK}
\affil[2]{\normalsize College of Computer Science and Engineering, University of Electronic Science and Technology of China, 611731, Chengdu, China}
\affil[3]{\normalsize Department of Electrical and Computer Engineering, East Lansing, MI 48824, USA}
\affil[4]{\normalsize Shenzhen Key Lab of Computational Intelligence, Department of Computer Science and Engineering, Southern University of Science and Technology, Shenzhen 518055, China}
\affil[5]{\normalsize CERCIA, School of Computer Science, University of Birmingham, Edgbaston, Birmingham, B15 2TT, UK}
\affil[$\ast$]{\normalsize Email: \texttt{k.li@exeter.ac.uk}}
\date{}

\begin{document}
\maketitle

\vspace{-3ex}
{\normalsize\textbf{Abstract: }The ultimate goal of multi-objective optimisation is to help a decision maker (DM) identify solution(s) of interest (SOI) achieving satisfactory trade-offs among multiple conflicting criteria. This can be realised by leveraging DM's preference information in evolutionary multi-objective optimisation (EMO). No consensus has been reached on the effectiveness brought by incorporating preference in EMO (either \textit{a priori} or \textit{interactively}) versus \textit{a posteriori} decision making after a complete run of an EMO algorithm. Bearing this consideration in mind, this paper $i$) provides a pragmatic overview of the existing developments of preference-based EMO; and $ii$) conducts a series of experiments to investigate the effectiveness brought by preference incorporation in EMO for approximating various SOI. In particular, the DM's preference information is elicited as a reference point, which represents her/his aspirations for different objectives. Experimental results demonstrate that preference incorporation in EMO does not always lead to a desirable approximation of SOI if the DM's preference information is not well utilised, nor does the DM elicit invalid preference information, which is not uncommon when encountering a black-box system. To a certain extent, this issue can be remedied through an interactive preference elicitation. Last but not the least, we find that a preference-based EMO algorithm is able to be generalised to approximate the whole PF given an appropriate setup of preference information.}
% in comparison with their non-preference-based baseline algorithms. In particular, the DM's preference information is elicited as a reference point, which represents her/his expectations on different objectives, in a priori manner. Experimental results demonstrate the usefulness of preference incorporation, especially for problems with many objectives. Moreover, we also find that the performance of different preference-based EMO algorithm can be largely influenced by their ways of how the preference information is used.}


%!Tex root=main.tex

\section{Introduction}
\label{sec:introduction}

It is not uncommon that real-world decision problems require solutions to simultaneously meet multiple objectives, known as multi-objective optimisation problems (MOPs). Note that these objectives are conflicting where an improvement in one objective can lead to a detriment of other objective(s). Hence, there does not exist a global optimum that optimises all objectives simultaneously. Instead, there exists a set of solutions representing the trade-offs among conflicting objectives. Generally speaking, a minimisation MOP considered in this paper is defined as follows:
\begin{equation}
\begin{array}{l l}
\mathrm{minimise} \quad \mathbf{F}(\mathbf{x})=(f_{1}(\mathbf{x}),\cdots,f_{m}(\mathbf{x}))^{T}\\
\mathrm{subject\ to} \quad \mathbf{x} \in\Omega
%\quad g_i(\mathbf{x})\geq a_i,\quad i=1,\cdots,q\\
%\mathrm{\ } \quad\quad\quad\quad\quad h_i(\mathbf{x})=b_i,\quad i=q+1,\cdots,\ell\\
%\mathrm{\ } \quad\quad\quad\quad\quad \mathbf{x} \in\Omega
\end{array},
\label{MOP}
\end{equation} 
where $\mathbf{x}=(x_1,\ldots,x_n)^T$ is a decision vector and $\mathbf{F}(\mathbf{x})$ is a objective vector. $\Omega=[x_i^L,x_i^U]^n\subseteq\mathbb{R}^n$ defines the search space. $\mathbf{F}: \Omega\rightarrow\mathbb{R}^m$ is the corresponding attainable set in the objective space $\mathbb{R}^m$. Without considering any preference information from a decision maker (DM), given two solutions $\mathbf{x}^1,\mathbf{x}^2\in\Omega$, $\mathbf{x}^1$ is said to dominate $\mathbf{x}^2$ if and only if $f_i(\mathbf{x}^1)\leq f_i(\mathbf{x}^2)$ for all $i\in\{1,\cdots,m\}$ and $\mathbf{F}(\mathbf{x}^1)\neq\mathbf{F}(\mathbf{x}^2)$. A solution $\mathbf{x}\in\Omega$ is said to be Pareto-optimal if and only if there is no solution $\mathbf{x}^\prime\in\Omega$ that dominates it. The set of all Pareto-optimal solutions is called the Pareto set (PS) and their corresponding objective vectors form the Pareto front (PF).

Due to the population-based property, evolutionary algorithms (EAs) have been widely recognised as a major approach for MO. Over the past three decades and beyond, many efforts have been dedicated to developing evolutionary multi-objective optimisation (EMO) algorithms, such as non-dominated sorting genetic algorithm II (NSGA-II)~\cite{DebAPM02}, indicator-based EA (IBEA)~\cite{ZitzlerK04} and multi-objective EA based on decomposition (MOEA/D)~\cite{ZhangL07}, to find a set of well-converged and well-diversified efficient solutions that approximate the whole PF. Nevertheless, the ultimate goal of MO is to help the DM identify a handful of representative solutions that meet at most her/his preferences. This inspires the requirement to incorporate the DM's preference information into MO -- techniques have been studied in the multi-criterion decision-making (MCDM) community over half a century. There are three classes of hybrid techniques considering the synergy of EMO and MCDM: \textit{a posteriori}, \textit{a priori}, and \textit{interactive}.

The traditional EMO follow the \textit{a posteriori} decision-making where a set of widely spread trade-off alternatives are obtained by an EMO algorithm before being presented to the DM. However, this not only increases the DM's workload, but also provides much irrelevant or even noisy information during the decision-making process. Due to the curse of dimensionality, the performance of EMO algorithms degenerate with the number of objectives~\cite{LiLTY15}. In addition, the number of points used to represent a PF grows exponentially with the number of objectives, thereby increasing the computational burden of an EMO algorithm. Besides, there is a severe cognitive obstacle for the DM to comprehend a high-dimensional PF.

If the preference information is elicited \textit{a priori}, it is used as a criterion to evaluate the fitness of a solution in the environmental selection and to drive the population towards the region(s) of interest (ROI) along a pre-defined \lq preferred\rq\ direction. In particular, the preference information can be represented as one or more reference points~\cite{BrankeKS01,Deb03,BrankeD05,LuqueSHCC09,ThieleMKL09}, reference directions~\cite{DebK07GECCO}, light beams~\cite{DebK07CEC} or value functions (VFs)~\cite{GreenwoodHD96}. Note that, in the \textit{a priori} approach, the DM only interact with the algorithm at the outset of an EMO process. It is controversial that the DM is able to faithfully represent her/his preference information before solving the MOP at hand.

%As for \textit{a priori} and \textit{interactive} approaches to hybridise EMO and MCDM, they all consider incorporating the DM's preference information into the search process. By doing so, computational efforts are expected to concentrate on the region of interest (ROI), thus having a better exploration therein. 
%\begin{itemize}
%    \item 
    %It is interesting to note that Greenwood et al~\cite{GreenwoodHD96}. derived a linear value function from a given ranking of a few alternatives to model the DM's preference information before optimisation. Thereafter, this linear value function is used as the fitness function in an EMO algorithm to guide the population towards the ROI. 
    %\item 
As for the \textit{interactive} preference elicitation, it enables the DM to progressively learn and understand the characteristics of the MOP at hand and adjust her/his elicited preference information. Consequently, solutions are gradually driven towards the ROI. In principle, many \textit{a priori} EMO approaches can be used in an interactive manner (e.g.~\cite{DebK07GECCO} and~\cite{DebK07CEC}). Specifically, in the first round, the DM can elicit certain preference information and it is used in an EMO algorithm to find a set of preferred non-dominated solutions. Thereafter, a few representative solutions will be presented to the DM. If these solutions are satisfactory, they will be used as the outputs and the iterative procedure terminates. Otherwise, the DM will adjust her/his preference information accordingly and it will be used in another EMO run. Alternatively, the DM can be involved to periodically provide her/his preference information as the EMO iterations are underway~\cite{PhelpsK03}. In particular, the preference information is progressively learned as VFs with the evolution of solutions. Since the DM gets a more frequent chance to provide new information, as discussed in~\cite{DebSKW10}, the DM may feel more in charge and more involved in the overall \textit{optimization-cum-decision-making} process.
%\end{itemize}

Although many efforts have been devoted to the synergy of EMO and MCDM, there is no systematic study, at least to the best of our knowledge, to investigate the \textit{pros and cons} brought by preference incorporation in EMO for approximating the ROI. This might be because although all preference-based EMO algorithms claim to approximate a ROI, the definition of the ROI is vague. In principle, it depends on the way how the DM elicits her/his preference information. For example, if the DM's preference information is elicited as a reference point, the ROI corresponds to a PF segment \lq close\rq\ to this reference point. On the other hand, if the DM's preference information is elicited as a VF learned from pair-wise comparisons made by the DM, it is difficult to define a specific location of the ROI on the PF. Instead, the preferred solutions are subjectively determined by the DM. Partially due to this reason, it is difficult to quantitatively evaluate the quality of preferred solutions obtained by various preference-based EMO algorithms under a unified framework.

In addition, although we criticised the ineffectiveness of \textit{a posteriori} decision-making process at the outset of this paper, there is no conclusive evidence to support the assertion that incorporating preference in EMO is superior to the traditional EMO for approximating solution(s) of interest (SOI). In particular, since the search process of a preference-based EMO algorithm is usually restricted to a certain region tentatively towards the ROI, it has a risk of losing population diversity and end up converging to an unexpected region.

Bearing the above mentioned considerations in mind, this paper empirically investigates the effectiveness of different algorithms, including both preference- and non-preference-based ones, for approximating various SOI. In particular, we assume that the DM elicits her/his preference information as a reference point $\mathbf{z}^r=(z_1^r,\cdots,z_m^r)^T$ where each component represents the DM's expected value on that objective. To have a quantitative comparison, we use our recently developed R-metrics~\cite{LiDY17} to evaluate the quality of obtained preferred solutions. In this paper, we aim to address the following five research questions (RQs) through our empirical studies.

\vspace{0.6em}
\noindent
\framebox{\parbox{\dimexpr\linewidth-2\fboxsep-2\fboxrule}{
\begin{enumerate}[$RQ1:$]
    \item \textit{Is preference incorporation in EMO really superior to traditional EMO for approximating SOI?}
    \item \textit{What is the most effective way to utilise the preference information in EMO for approximating SOI?}
%            Different preference-based EMO algorithms have distinct ways of using the preference information, how do these differences reflect on the performance for approximating the ROI?}
    \item \textit{How does the location of a reference point influence the performance of a preference-based EMO algorithm for approximating SOI?}
    \item \textit{If the DM's preference information is set in an interactive manner according to the evolution status, how does it influence the results?}
    \item \textit{Is that possible to generalise the preference-based EMO to approximate the whole PF rather than merely a partial region?}
\end{enumerate}
}}
\vspace{0.5em}

In the rest of this paper, \pref{sec:review} provides a pragmatic review of the current developments of preference-based EMO. \pref{sec:setup} describes the methodologies that we used to setup the experiments, including algorithms, benchmark problems, different reference point settings, performance metrics. \pref{sec:experiments} presents and analyses the experimental results in accordance with the RQs. At the end, \pref{sec:conclusions} concludes this paper and provides some future directions.

%!Tex root=main.tex

\section{Literature Review}
\label{sec:review}

\begin{figure}[htbp]
    \centering
    \includegraphics[width=\linewidth]{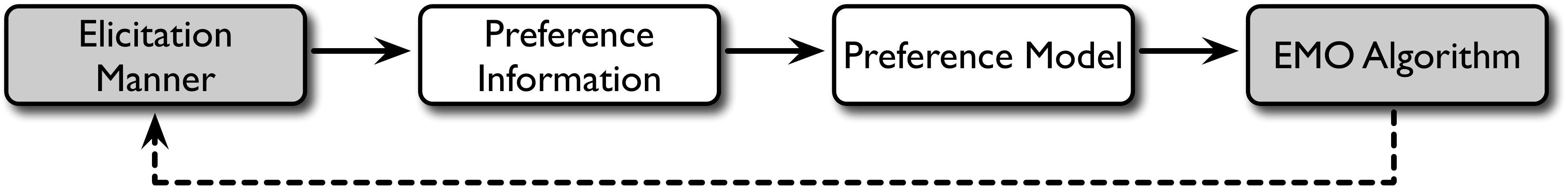}
    \caption{Flowchart of a PBEMO process.}
    \label{fig:PEMO}
\end{figure}

As introduced in~\pref{sec:introduction}, there is a growing trend of incorporating the DM's preference information into EMO~\cite{Coello00,RachmawatiS06,PurshouseDMMW14,BechikhKSG15} to approximate her/his preferred Pareto-optimal solutions in the past decades. Generally speaking, a preference-based EMO (PBEMO) process can be broken down into four essential components as shown in~\pref{fig:PEMO}.

\begin{itemize}
    \item The elicitation manner decides when to ask the DM to elicit her/his preference information. There are three different elicitation manners: \textit{a posteriori} (i.e. after a complete run of an EMO algorithm), \textit{a priori} (i.e. before running an EMO algorithm) and \textit{interactive} (i.e. during the running of an EMO algorithm).

    \item The preference information is the way of how does the DM express her/his preference. Perhaps the most straightforward one is a reference point, as known as an aspiration level vector, which represents the expected value the DM wants to achieve. The other one is through holistic comparisons which can be based on either the comparisons on solutions or objective functions. As for this latter one, it can be implemented as pairwise comparisons and a qualitative classification of solutions, etc. Whilst the comparison on objective functions can be realised by assigning weights to different objectives, a redefinition of trade-off relation and a classification of objective functions.

    \item The preference model is the way how the preference information elicited by the DM can be used in an EMO algorithm. In the literature, VF, dominance relation and decision rules~\cite{SlowinskiGM02} are the most popular choices. In particular, a VF is a scalar function of all objectives which evaluates solutions quantitatively. Dominance relation describes the DM’s preference in the form of the relation of a pair of solutions. Decision rules model the DM's preference as a set of \lq\texttt{IF-THEN}\rq\ rules.

    \item An EMO algorithm is the search engine that iteratively approximates SOI according to the preference model. In principle, any EMO framework (i.e. dominance-\cite{DebK07GECCO,DebK07CEC,LuqueSHCC09}, indicator-~\cite{ThieleMKL09} and decomposition-based frameworks~\cite{WickramasingheL08,GongLZJZ11,LiCMY18,LiCSY18}) can be used at this stage.
\end{itemize}

Note that the PBEMO process shown in~\pref{fig:PEMO} is a closed-loop system when using an \textit{interactive} elicitation manner. Otherwise, it is a one-off process. In particular, the EMO algorithm is the starting point of this process when the elicitation manner is \textit{a posteriori}. On the other hand, it is the ending point when \textit{a priori} elicitation manner is used. In the following paragraphs, we will provide an overview on the current development of PBEMO mainly according to the elicitation manner, intertwined with the preference information and the preference model.

\subsection{A Priori Elicitation Manner}
\label{sec:a_priori}

\subsubsection{Using Reference Point as Preference Infromation}
%To incorporate the DM's preference information into the search process beforehand, the most widely used way to express and model the DM's preference information is to specify a reference point.
This is the most widely used way to express and model the DM's preference information. The first attempt along this line is from Fonseca and Fleming~\cite{FonsecaF98} who suggested to model the DM's preference as a goal that indicates desired levels of performance in each objective dimension. Afterwards, the reference point(s) were used in various ways to guide the EMO process towards the ROI. For example, in~\cite{DebSBC06,DebK07CEC} and~\cite{DebK07}, Deb et al. used the Euclidean distance to the reference point(s) as a second criterion (additional to the Pareto dominance) to evaluate the fitness of a solution. In particular, solutions closer to the reference point(s) have a higher priority to survive. Based on the similar merit, in~\cite{WickramasingheL08} and~\cite{AllmendingerLB08}, the reference point is used to help select the leader swarm in the multi-objective particle swarm optimisation algorithm. In~\cite{ThieleMKL09}, to consider DM's preference information, Thiele et al. made a simple modification on IBEA by incorporating the achievement scalarising function (ASF) into a binary indicator.

Furthermore, the relative position with respect to the reference point can be used to define a new dominance relation as well. For example, Molina et al.~\cite{MolinaSDCC09} suggested the g-dominance where solutions satisfying either all or none aspiration levels are preferred over those satisfying some aspiration levels. Said et al.~\cite{SaidBG10} developed the r-dominance, where non-dominated solutions, according to the Pareto dominance relation, can be differentiated by their weighted Euclidean distances towards the reference point.

Instead of being directly used to guide solutions towards the ROI, the reference point can also be used to change the distribution of weight vectors, which are the core design components in the emerging decomposition-based EMO algorithms, according to the DM's preference information, e.g.~\cite{MohammadiOLD14,Yu2015,Ma2015,LiCMY18}. In~\cite{NarukawaSTOSI16}, Narukawa et al. proposed an interesting preference-based NSGA-II where the DM's preference information is expressed as Gaussian functions on a hyperplane. In addition, reference points are also core components of $R2$ indicator, a set-based performance indicator. In~\cite{TrautmannWB13} and~\cite{WagnerTB13}, $R2$ indicator is modified to consider the DM's preference information.

Comparing to the other preference modelling tools, reference point is relatively intuitive to represent the DM's preference information. Without a demanding effort, the DM is able to guide the search towards the ROI directly or interactively even when encountering a large number of objectives. Recently, the first author and his collaborators developed a systematic way to evaluate and compare the performance of preference-based EMO algorithms using reference points for approximating the ROI~\cite{LiDY17}. This work lays the foundation to rigorously evaluate and compare different preference-based EMO algorithms by using reference point(s).

\subsubsection{Using Weights as Preference Information}

%The second \textit{a priori} preference elicitation method is to assign weights to different objectives according to their relative importance. 

Its basic idea is to assign weights to different objectives according to their relative importance. For example, Deb~\cite{Deb03} developed a modified fitness sharing mechanism, by using a weighted Euclidean distance, to bias the population distribution. Branke et al.~\cite{BrankeKS01} proposed a modified dominance principle where the trade-off among two objectives is directly specified by the DM, e.g. a gain/degradation of an objective by one unit will lead to a corresponding degradation/gain in the other objective. In~\cite{BrankeD05}, Branke and Deb developed a linearly weighted utility function that projects solutions to a hyperplane before evaluating the crowding distance in NSGA-II. In~\cite{ZitzlerBT06}, Zitzler et al. showed how to use a weight distribution function on the objective space to incorporate preference information into Hypervolume-based EMO algorithms. Based on the same merit, Friedrich et al.~\cite{FriedrichKN13} generalised this idea to two dominance-based EMO algorithms NSGA-II and SPEA2~\cite{ZitzlerLT01}.

It is worth noting that the weight-based methods become ineffective when facing a large number of objectives. Because it is difficult to neither specify the weights nor verify the quality of the biased approximation. Moreover, it is unintuitive and challenging for the DM to steer the search process towards the ROI via the weighting scheme. In addition, the weight-based methods are unable to approximate multiple ROIs and control the extent of the ROI.

\subsubsection{Using Desirability Function (DF) as Preference Information}

DF~\cite{WagnerT10} aims to map each individual objective into a desirability with a value bounded within the range $[0,1]$. Through this mapping, values of different objectives become comparable. Moreover, DFs are also able to prevent a biased distribution of solutions caused by badly scaled objectives. Afterwards, DFs are integrated with a popular indicator-based EMO algorithm, i.e. SMS-EMOA~\cite{BeumeNE07}, to approximate the ROI. Note that the calculation of Hypervolume (HV) is based on the DFs instead of the original objective functions.

\subsection{Interactive Elicitation Manner}
\label{sec:interactive_literature}

In fact, almost all methods developed under \textit{a priori} preference elicitation setting can be applied in an \textit{interactive} manner. For example, the DM can periodically adjust the reference point to progressively guide the population towards the ROI. In the following paragraphs, we will overview some representative developments on the interactive MO.

\subsubsection{Using Fuzzy Function as Preference Information}

By classifying the relative importance of objectives into different grades, Cvetkovi\'c and Parmee~\cite{ParmeeC02} developed a fuzzy preference relation that translates the pairwise comparisons among objectives into a weighted-dominance relation. In~\cite{JinS02}, Jin and Sendhoff developed a method to convert the DM's fuzzy preference information into weight intervals through pairwise comparisons on objectives. Shen et al.~\cite{ShenGCH10} proposed an interactive EMO algorithm based on fuzzy logic. In particular, after running the EMO algorithm for several generations, the DM is asked to specify the relative importance between pairs of objectives via linguistic terms. Thereafter, a new fitness function is defined according to a \lq strength superior\rq\ relation derived from a fuzzy inference system.

\subsubsection{Using Value Function (VF) as Preference Information}

%Pairwise comparisons among a set of selected candidate solutions is another popular way to collect the DM's preference information in an interactive manner. For example, 
As a pioneer along this line, Phelps and K\"oksalan~\cite{PhelpsK03} proposed an interactive evolutionary meta-heuristic algorithm that translates the DM's pairwise comparisons of solutions into a linear programming problem. In particular, its optimal solution is the weights of an estimated VF in the form of a weighted sum whilst the estimated VF is used as the fitness function of the evolutionary meta-heuristic algorithm. In~\cite{BattitiP10}, Battiti and Passerini developed a progressively interactive EMO approach that uses learning-to-rank method to estimate the parameters of a polynomial VF. Afterwards, the derived VF is used to modify the Pareto dominance to compare solutions. In~\cite{DebSKW10}, Deb et al. developed an interactive EMO algorithm that progressively learns an approximated VF by asking the DM to compare a set of solutions in a pairwise manner. In~\cite{BrankeGSZ15}, Branke et al. proposed to use robust ordinal regression to learn a representative additive monotonic VF compatible with the DM's preference information. Thereafter, the VF is used to replace the crowding distance calculation in NSGA-II. In~\cite{PedroT14}, Pedro and Takahashi proposed to use a Kendall-tau distance to evaluate the accuracy of the approximated VF learned by a radial basis function network. If the approximated VF is satisfactory, it is used to dynamically change the calculation of the crowding distance in NSGA-II to manipulate the density of solutions in a population. Instead of modelling the DM's preference information as VFs, Greco et al.~\cite{GrecoMS11} proposed to use decision rules to implement the preference modelling.

In~\cite{MiettinenM00}, Miettinen and M{\"{a}}kel{\"{a}} developed an interactive multi-objective optimisation system called WWW-NIMBUS that allows the DM to classify objectives into up to five classes so as to find a more desirable solution. During the search process, the original MOP is transformed into a constrained single-objective optimisation problem by combining a weighted distance metric with an ASF. Later, Miettinen et al.~\cite{MiettinenERL10} proposed the NAUTILUS method that starts from the nadir point and improves all objectives simultaneously in an interactive manner. In particular, the DM is able to specify either the frequency of interaction or the percentages of which (s)he would like to improve at each objective. Note that both WWW-NIMBUS and NAUTILUS use the classic mathematical programming techniques as the search engine. In~\cite{SindhyaRM11}, Sindhya et al. proposed to use EA to search for SOI under the framework of NAUTILUS.

In~\cite{Yang99,YangL02,LuqueYW09}, Yang et al. proposed GRIST method that estimates the gradient of an underlying VF by using the indifference trade-offs provided by the DM in an interactive manner. Thereafter, the gradient is projected onto the tangent hyperplane of the PF so that the search process can be guided towards the direction along which the DM's utility can be improved. Recently, Chen et al.~\cite{ChenXC17} applied the GRIST method in the context of EA to improve the versatility of the GRIST method for solving problems without nice mathematical properties such as convexity and differentiability.

\subsubsection{Using Holistic Comparisons as Preference Information}

Asking the DM to periodically select the most preferred solution from a set of candidates is another alternative way to represent the DM's preference information. For example, Folwer et al.~\cite{FowlerGKKMW10} proposed to use the best and the worst solutions specified by the DM to construct convex preference cones. Thereafter, a cone dominance relation is defined to rank the population. In~\cite{SinhaKWD10}, Sinha et al. proposed a progressively interactive EMO algorithm that asks the DM to select the most preferred solution from an archive. The collected preference information is used to build polyhedral cones for modifying Pareto dominance relation. In~\cite{KoksalanK10}, K\"oksalan and Karahan proposed an interactive version of territory defining EA~\cite{KarahanK10} to consider the DM's preference information in the loop. In particular, a territory is defined around each individual and the favourable weights of the best solution selected by the DM are identified to determine a new preferred weight region.

In~\cite{GongLZJZ11}, Gong et al. proposed an interactive MOEA/D where the weight vector of the selected best solution is used to renew the preferred weight region. In particular, this region is a hyper-sphere with the preferred weight vector being the centre. Recently, the first author and his collaborators~\cite{LiCSY18} proposed a systematic framework for incorporating the DM's preference information into the decomposition-based EMO algorithms. More specifically, it periodically asks the DM to score a couple of selected solutions according to their satisfaction to the DM's preference information. Based on the scoring results, a radial basis function network is trained to predict the fitness of solutions in the next several generations. Moreover, the fitness of solutions directly represent the priority of weight vectors. In other words, the best solution is associated with the most promising weight vector, so on and so forth. The other weight vectors are moved towards those selected promising weight vectors to represent the DM's preference information.

\subsection{A Posteriori Elicitation Manner}
\label{sec:interactive} 

In the \textit{a posteriori} scenario, the DM has no chance to modify the existing trade-off alternatives obtained by an EMO algorithm. Instead, the \textit{a posteriori} methods mainly aim to shortlist solutions that might be interested by the DM to support the decision-making process. The most popular one is to identify the knee points of which a small improvement in one objective can lead to a large deterioration in other objectives~\cite{DebG11}. For example, Bhattacharjee et al.~\cite{BhattacharjeeSR17} developed a method that recursively uses the expected marginal utility measure to identify the SOI. Moreover, this method is also able to characterise the nature of those selected solutions (either internal or peripheral) through a set of systematically generated reference directions. Besides knee points, solutions lying on the edge of the approximated PF is useful for the DM to understand some important characteristics of the PF, e.g. its shape and boundary. In~\cite{EversonWF14}, Everson et al. proposed four definitions of edge points and examined their relations under a many-objective setting.

Different from the knee and edge points, subset selection is another alternative to find a pre-specified number of solutions that best represent the characteristics of the original PF. To this end, researchers (e.g.~\cite{SinghBR18} and~\cite{IshibuchiISN17}) mainly aim to efficiently choose a limited number of representative solutions that achieve an inverted generational distance (IGD)~\cite{BosmanT03} minimisation or a HV~\cite{ZitzlerT99} maximisation.

%!TeX root=main.tex

\section{Experimental Setup}
\label{sec:setup}

This section introduces the setup of our experiments, including the basic mechanisms of the selected traditional and preference-based EMO algorithms; the characteristics of the benchmark problems; the settings of reference points that represent various DM's preference information; and the performance metrics used to evaluate the quality of solution sets for approximating the ROI. More detailed settings can be found in Section 1 of the supplementary document of this paper\footnote{\url{http://cola-laboratory.github.io/publications/supp-case.pdf}}.

\vspace{-0.7em}

\subsection{Peer Algorithms}
\label{sec:algorithms}

It is well known that there are three major frameworks (i.e. dominance-, indicator- and decomposition-based frameworks) in the EMO literature~\cite{LiZZL09,LiZLZL09,LiFK11,LiKWCR12,CaoKWL12,LiKWTM13,LiWKC13,LiK14,CaoKWL14,LiZKLW14,LiFKZ14,CaoKWLLK15,WuKZLWL15,LiDZK15,LiKZD15,LiKD15,LiDZZ17,WuKJLZ17,WuLKZZ17,LiDY18,ChenLY18,WuLKZ18,ChenLBY18,LiCFY19,WuLKZZ19,LiCSY19,Li19,BillingsleyLMMG19,ZouJYZZL19}. To study the effectiveness of preference incorporation in EMO, peer algorithms are chosen in accordance to this categorisation. In particular, we choose three iconic EMO algorithms, i.e. NSGA-III~\cite{DebJ14}, IBEA~\cite{ZitzlerK04} and MOEA/D~\cite{ZhangL07} without considering the DM's preference information. Note that all of them are scalable to handle problems with more than three objectives. In addition, we choose six widely used preference-based EMO algorithms, i.e. R-NSGA-II~\cite{DebSBC06}, r-NSGA-II~\cite{SaidBG10}, g-NSGA-II~\cite{LuqueSHCC09}, PBEA~\cite{ThieleMKL09}, RMEAD2~\cite{MohammadiOLD14} and MOEA/D-NUMS~\cite{LiCMY18} in our experiments. Note that all preference-based EMO algorithms use reference point(s) to represent the DM's preference information. Their differences mainly lie in the way of how to utilise the preference information to drive the search process. The following paragraphs briefly introduce the mechanisms of these selected peer algorithms whilst interested readers can find more details from their original papers.

\subsubsection{Traditional EMO Algorithms}
\label{sec:traditional}

\begin{itemize}
    \item\underline{NSGA-III}: it is an extension of NSGA-II where the mixed population of parents and offspring is first divided into several non-dominated fronts by using the fast non-dominated sorting procedure. Afterwards, solutions in the first several fronts have a higher priority to survive to the next generation. In particular, the exceeded solutions are trimmed according to the local density of a subregion specified by one of the evenly sampled weight vectors.

    \item\underline{IBEA}: it transfers an MOP into a single-objective optimisation problem that optimises a binary performance indicator, e.g. the binary additive $\epsilon$-indicator as:   
%    it defines the optimisation goal as a binary performance indicator, e.g. the binary additive $\epsilon$-indicator as:
    \begin{equation}
    \begin{split}
        I_{\epsilon^+}(A,B)=\min_{\epsilon}\Big\{\forall\mathbf{x}^2\in B, \exists\mathbf{x}^1\in A: f_i(\mathbf{x}^1)-\epsilon\\
        \leq f_i(\mathbf{x}^2), i\in\{1,\cdots,m\}\Big\}
    \end{split},
    \end{equation}
    Then, this indicator is directly used to assign the fitness value to a solution $\mathbf{x}$ in the current population $P$:
    \begin{equation}
        F(\mathbf{x})=\sum_{\mathbf{x}'\in P\setminus\{\mathbf{x}\}}-e^{-I_{\epsilon^+}(\{\mathbf{x}'\},\{\mathbf{x}\})/\kappa},
    \end{equation}
    %By doing so, an MOP is naturally transformed into a single-objective optimisation problem.

    \item\underline{MOEA/D}: its basic idea is to decompose the original MOP into several subproblems, either as a single-objective scalarising function or a simplified MOP. Then, a population-based technique is used to solve these subproblems in a collaborative manner. In particular, this paper chooses the widely used inverted Tchebycheff function as the subproblm formulation:
\begin{equation}
\begin{array}{l l}
\mathrm{minimise}\quad g^{tch}(\mathbf{x}|\mathbf{w},\mathbf{z}^{\ast})=\max\limits_{i=1,\cdots,m}\{|f_i(\mathbf{x})-z_{i}^{\ast}|/w_i\}\\
\mathrm{subject\ to}\quad \mathbf{x}\in\mathbf{\Omega}
\end{array},
\label{eq:TCH}
\end{equation}
For convenience, we allow $w_i=0$ in setting $\mathbf{w}$, but replace $w_i=0$ by $w_i=10^{-6}$ in \pref{eq:TCH}. %The direction vector for this subproblem is $\mathbf{w}=(w_1,\ldots,w_m)^T$.
\end{itemize}

\subsubsection{Preference-based EMO Algorithms}
\label{sec:preference}

\begin{itemize}
    \item\underline{R-NSGA-II}: it uses the weighted distance between a solution $\mathbf{x}$ (belonging to the last acceptable non-dominated front) and $\mathbf{z}^r$ to replace the crowding distance of NSGA-II. In particular, the weighted distance is calculated as: 
    \begin{equation}
        Dist(\mathbf{x}, \mathbf{z}^r) = \sqrt{\sum_{i=1}^{m}w_i \left(\frac{f_i(\mathbf{x}) - f_i(\mathbf{z}^r)}{f_i^{\max} - f_i^{\min}}\right)^2},
    \label{eq:weighted_distance}
    \end{equation}
    where $\sum_{i=1}^mw_i=1$ and $w_i\in[0,1]$. $f_i^{\max}$ and $f_i^{\min}$ are respectively the maximum and minimum at the $i$-th objective. Furthermore, R-NSGA-II uses an $\epsilon$-clearing strategy to avoid over-crowdedness within a local niche.

    \item\underline{r-NSGA-II}: it defines a new dominance relation, called r-dominance, to incorporate the DM's preference information in NSGA-II. Specifically, given two solutions $\mathbf{x}^1$ and $\mathbf{x}^2$, $\mathbf{x}^1$ is said to r-dominate $\mathbf{x}^2$ if $\mathbf{x}^1$ dominates $\mathbf{x}^2$; or $\mathbf{x}^1$ and $\mathbf{x}^2$ are non-dominated according to the Pareto dominance, and $\overline{Dist}(\mathbf{x}^1,\mathbf{x}^2,\mathbf{z}^r)<-\delta$, where: 
    \begin{equation}
        \overline{Dist}(\mathbf{x}^1,\mathbf{x}^2,\mathbf{z}^r)=\frac{Dist(\mathbf{x}^1,\mathbf{z}^r) - Dist(\mathbf{x}^2,\mathbf{z}^r)}{Dist_{\max} - Dist_{\min}},
    \end{equation}
    where $Dist_{\max}$ and $Dist_{\min}$ are respectively the maximum and minimum of $Dist(\mathbf{x},\mathbf{z}^r)$ in the current population. $\delta\in[0,1]$ is used to control the extent of the approximated ROI.

\item\underline{g-NSGA-II}: it defines a new dominance relationship called g-dominance in NSGA-II. Given $\mathbf{z}^r$, solutions dominated by or dominate $\mathbf{z}^r$ are more preferable than those non-dominated ones.

%\pref{fig:g-dominance} gives a simple illustration of the basic idea of g-dominance. Given the DM supplied reference point $\mathbf{z}^r$, solutions located in the region denoted as Flag 1 are more preferable than those located in the region denoted Flag 0.
%        \begin{figure}[htbp]
%        	\centering
%            \includegraphics[width=.35\linewidth]{figs/g-dominance.pdf}
%            \caption{Illustration of g-dominance with respect to $\mathbf{z}^r$.}
%            \label{fig:g-dominance}
%        \end{figure}
    \item\underline{PBEA}: it integrates the DM's preference information into IBEA by modifying its additive $\epsilon$-indicator as follows:
    \begin{equation}
        I_p(\mathbf{x}^1,\mathbf{x}^2) = I_{\epsilon}(\mathbf{x}^1,\mathbf{x}^2) / (s(\mathbf{x}^1) + \sigma - \min_{\mathbf{x}^2 \in P}[s(\mathbf{x}^2)]),
    \end{equation}
    where $P$ is the current population, $\sigma>0$ controls the importance of different solutions with respect to $\mathbf{z}^r$. In particular, the smaller the $\sigma$ is, the more solutions near $\mathbf{z}^r$ are favoured, $s(\mathbf{x})$ is the augmented Tchebycheff ASF:
    \begin{equation}
        s(\mathbf{x})=\max_{i=1,\cdots,m}w_i(f_i(\mathbf{x})-z_i^r)+\rho\sum_{i=1}^m(f_i(\mathbf{x})-z_i^r),
    \end{equation}
    $\rho$ is a small augmentation coefficient. %and $\sigma>0$ is in general able control the importance of different solutions with respect to $\mathbf{z}^r$. In particular, the smaller the $\sigma$ is, the more solutions near $\mathbf{z}^r$ are favoured.
	\item\underline{RMEAD2}: it is a variant of MOEA/D where the DM's preference information is used to generate a set of weight vectors biased towards the DM supplied reference point. To this end, it gradually re-samples new weight vectors, according to a uniform distribution, in the vicinity of the solution with respect to the weight vector closest to $\mathbf{z}^r$.	
	\item\underline{MOEA/D-NUMS}: it uses a closed-form non-uniform mapping scheme to transform the originally evenly distributed weight vectors on a canonical simplex into new positions close to $\mathbf{z}^r$. Thereafter, the transformed weight vectors are used in MOEA/D or any other decomposition-based EMO algorithm to steer the search process towards the ROI either directly or interactively.
\end{itemize}

\vspace{-1em}

\subsection{Test Problems}
\label{sec:testproblems}

In this paper, we consider test problems chosen from two most popular benchmark suites, i.e. ZDT~\cite{ZitzlerDT00} and DTLZ~\cite{DebTLZ05}. The test problems therein are with continuous variables and have various PF shapes (e.g. linear, convex, concave and disconnected) and search space properties. ZDT problems have only two objectives, whilst the number of objectives of DTLZ problems varies from 3 to 10. The number of variables are set as recommended in~\cite{LiDZK15}.

\vspace{-1em}

\subsection{Settings of Reference Points}
\label{sec:refpoints}

In our experiments, we consider two types of reference point settings. One is called a \lq balanced\rq\ setting where the reference point is placed at the centre region of the PF; whilst the other is called an \lq extreme\rq\ setting where the reference point is placed at the vicinity of an extreme of the PF. For each case, we set three reference points, i.e. $\mathbf{z}^r_\texttt{P}$ on the PF, $\mathbf{z}^r_\texttt{I}$ in the infeasible region and $\mathbf{z}^r_\texttt{F}$ in the feasible region. %In particular, for the \lq balanced\rq\ reference point, $\mathbf{z}^r_\texttt{IF}$ and $\mathbf{z}^r_\texttt{F}$ is separately set as $\mathbf{z}^r_\texttt{PF}\pm 0.1\times\mathbf{I}$; while for those \lq extreme\rq\ case, they are set as $\mathbf{z}^r_\texttt{PF}$, except the $j$-th objective is separately set as $z^r_{\texttt{PF},j}\pm 0.1$ where $j=\argmax_{i\in\{1,\cdots,m\}}z^r_{\texttt{PF},i}$. \pref{fig:refpoint} gives an example of the \lq balanced\rq\ and \lq extreme\rq\ settings in a two-objective scenario. 
%\begin{figure}[htbp]
%    \centering
%    \includegraphics[width=.4\linewidth]{figs/refpoint.pdf}
%    \caption{Illustration of different reference points for the \lq\lq balanced\rq\rq\ setting.}
%    \label{fig:refpoint}
%\end{figure}

\subsection{Preference Metrics}
\label{sec:Rmetrics}

To have a quantitative comparison, we consider two levels of assessments. The first one is about the approximation accuracy. Given the DM supplied reference point $\mathbf{z}^r$, the approximation accuracy achieved by a solution set $P$ is evaluated as:
\begin{equation}
    \mathbb{E}(P)=\min_{\mathbf{x}\in P}\Big\{\max_{i=1,\cdots,m}(f_i(\mathbf{x})-z_i^r)/w_i\Big\},
\end{equation}
The smaller the $\mathbb{E}(P)$ is, the better $P$ is for approximating the DM most preferred solution. In particular, we set $w_i=\frac{1}{m}$ in our experiments given that all objectives are assumed to be of an equal importance.

The second assessment is based on our recently proposed R-metrics~\cite{LiDY17}, i.e. R-IGD and R-HV. They are used to evaluate the quality of an approximation set for approximating the ROI with respect to the DM supplied reference point. The basic idea of R-metric evaluation is to pre-process the approximation sets found by different algorithms before using the IGD and HV for performance assessment. More details related to the R-metric calculation can be found in Section 2 of the supplementary document.

In the experiments, each algorithm is performed 31 independent runs. We keep a record of the median and the interquartile range of metric values obtained for different test problems with various reference point settings. The corresponding data are gathered in Tables 6 to 35 in Section 3 of the supplementary document. In particular, the best metric values are highlighted in bold face with a grey background. To have a statistically sound conclusion, we use the Wilcoxon signed-rank test at a 0.05 significance level to validate the statistical significance of the best median metric values. In addition, we keep a record of the ranks, with respect to the performance metrics, achieved by different algorithms on each test case. These are visualised as the heat maps shown in Figs.~\ref{fig:EP_BAL} to~\ref{fig:RHV_EXT}.

%!TeX root=main.tex

\section{Empirical Results and Analysis}
\label{sec:experiments}

Due to the massive amount of data collected in our experiments, it will be messy if we pour all results in this paper. Instead, it is more plausible that we focus on some important observations contingent upon the RQs posed in~\pref{sec:introduction}. Whilst the complete results, including performance metric values and plots of population distribution are put in the supplementary document of this paper.
% and trajectories of R-HV and R-IGD values versus the number of generations, 

\vspace{-1em}

\subsection{Performance Comparisons of Preference and Non-Preference-Based EMO Algorithms}
\label{sec:response_RQ1}

\begin{figure*}[htbp]
    \centering
    \includegraphics[width=1.0\linewidth]{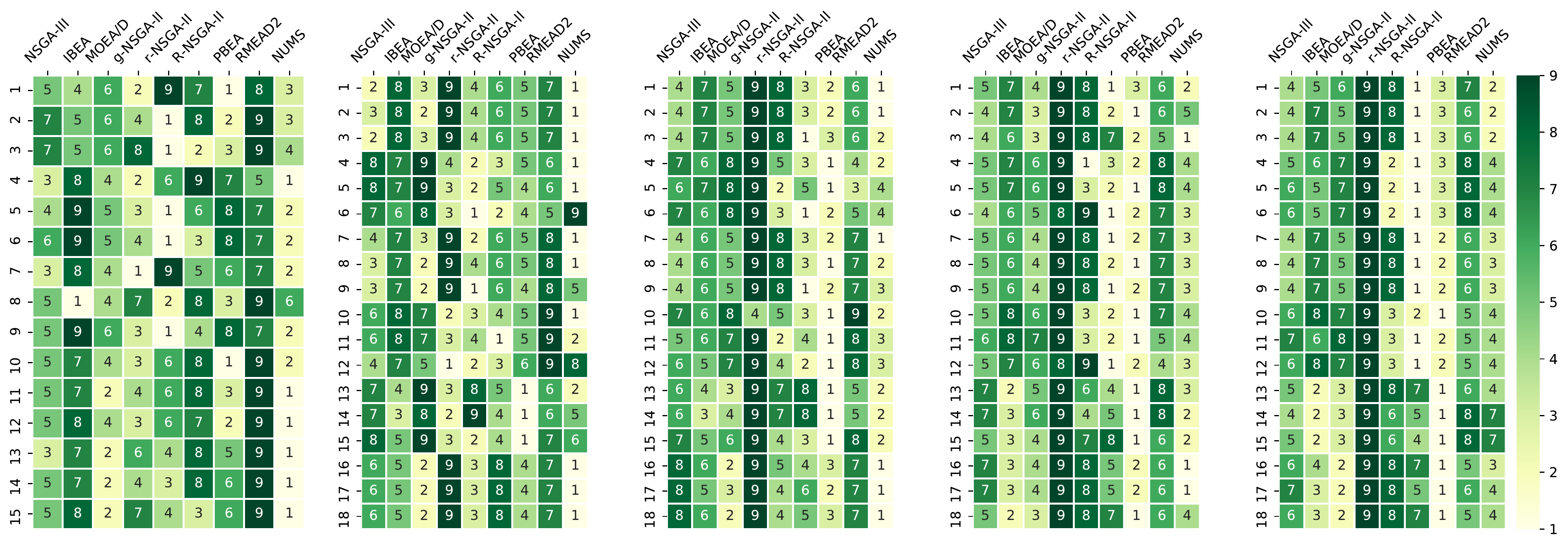}
    \caption{Heat maps of the ranks of $\mathbb{E}(P)$ obtained by different algorithms for the \lq balanced\rq\ reference point settings. The subplots, from left to right, are ZDT problems and DTLZ problems with 3 to 10 objectives, respectively.}
    \label{fig:EP_BAL}
\end{figure*}

\begin{figure*}[htbp]
    \centering
    \includegraphics[width=1.0\linewidth]{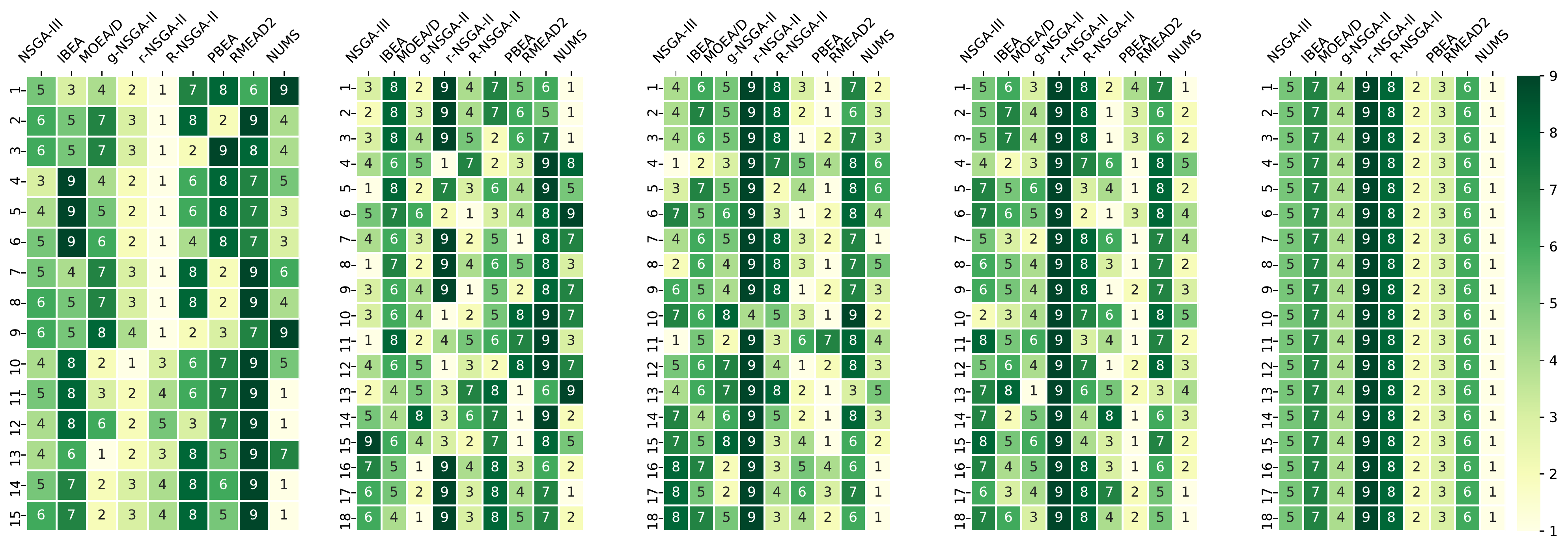}
    \caption{Heat maps of the ranks of $\mathbb{E}(P)$ obtained by different algorithms for the \lq extreme\rq\ reference point settings. The subplots, from left to right, are ZDT problems and DTLZ problems with 3 to 10 objectives, respectively.}
    \label{fig:EP_EXT}
\end{figure*}

\begin{figure*}[htbp]
    \centering
    \includegraphics[width=1.0\linewidth]{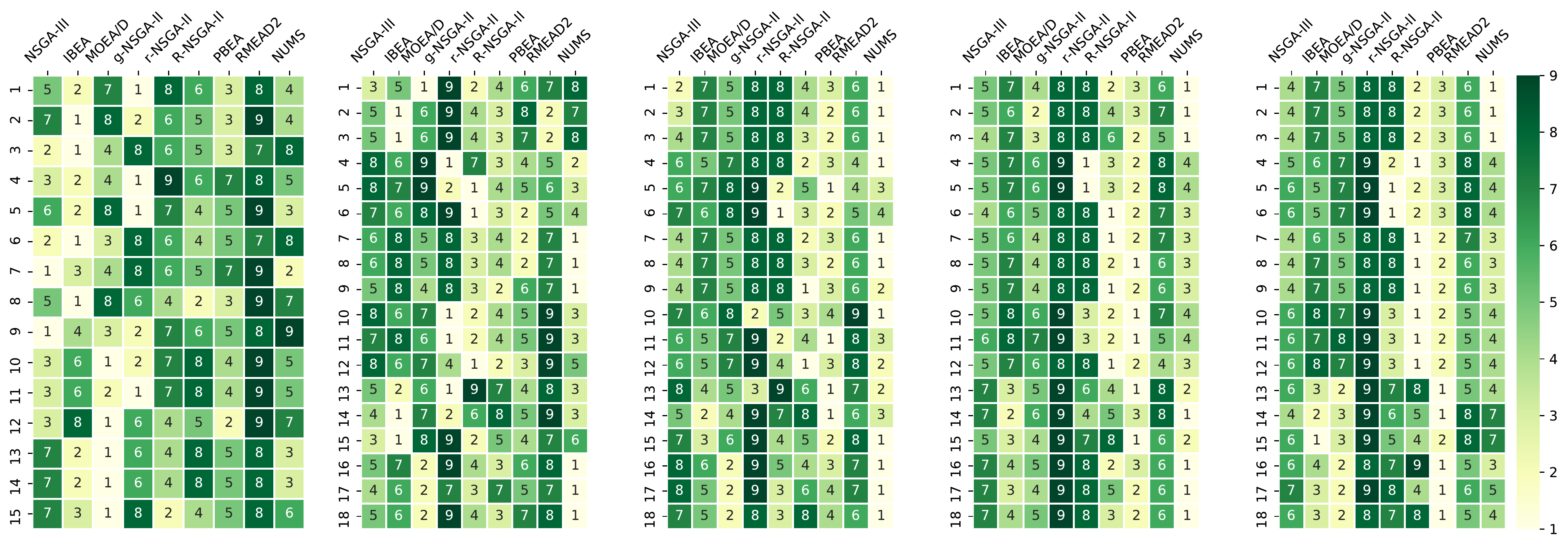}
    \caption{Heat maps of the ranks of R-IGD values obtained by different algorithms for the \lq balanced\rq\ reference point settings. The subplots, from left to right, are ZDT problems and DTLZ problems with 3 to 10 objectives, respectively.}
    \label{fig:RIGD_BAL}
\end{figure*}

\begin{figure*}[htbp]
    \centering
    \includegraphics[width=1.0\linewidth]{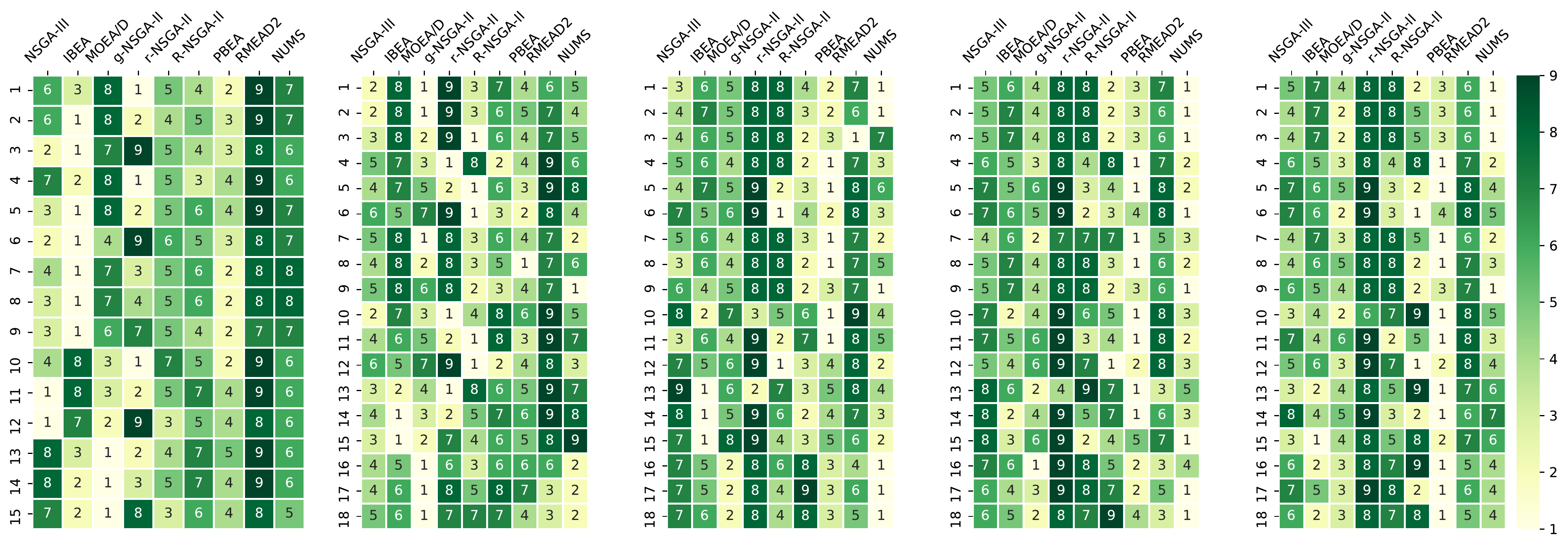}
    \caption{Heat maps of the ranks of R-IGD values obtained by different algorithms for the \lq extreme\rq\ reference point settings. The subplots, from left to right, are ZDT problems and DTLZ problems with 3 to 10 objectives, respectively.}
    \label{fig:RIGD_EXT}
\end{figure*}

\begin{figure*}[htbp]
    \centering
    \includegraphics[width=1.0\linewidth]{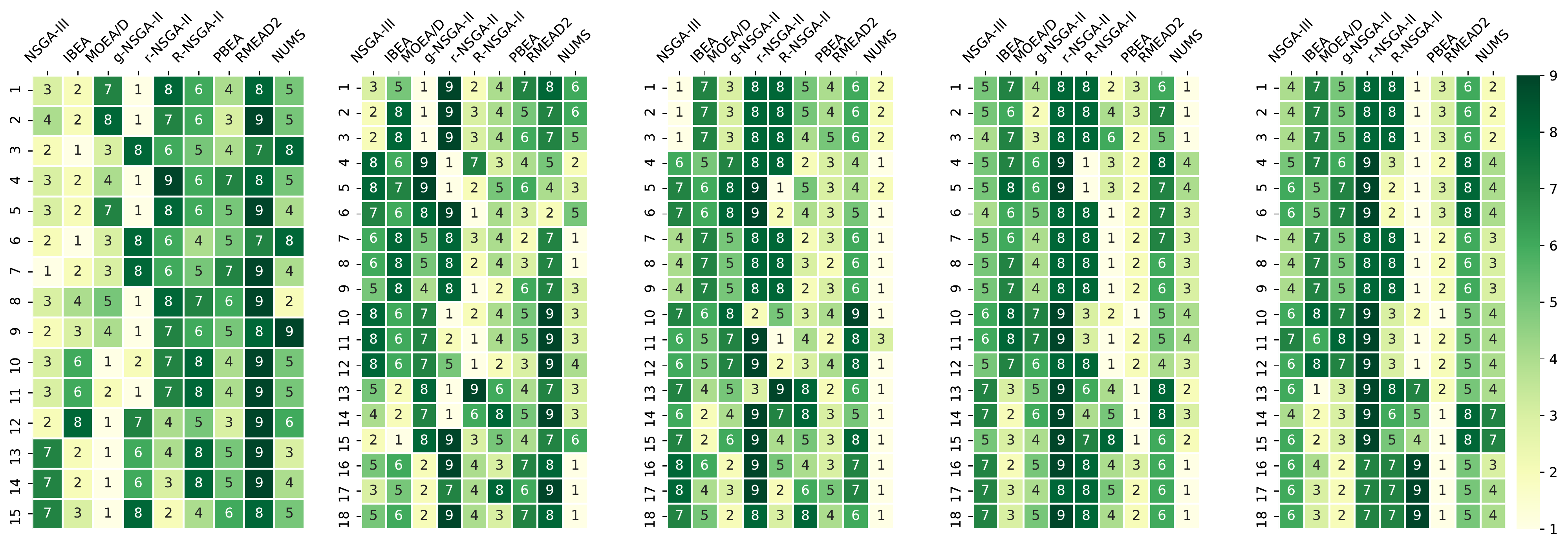}
    \caption{Heat maps of the ranks of R-HV values obtained by different algorithms for the \lq balanced\rq\ reference point settings. The subplots, from left to right, are ZDT problems and DTLZ problems with 3 to 10 objectives, respectively.}
    \label{fig:RHV_BAL}
\end{figure*}

\begin{figure*}[htbp]
    \centering
    \includegraphics[width=1.0\linewidth]{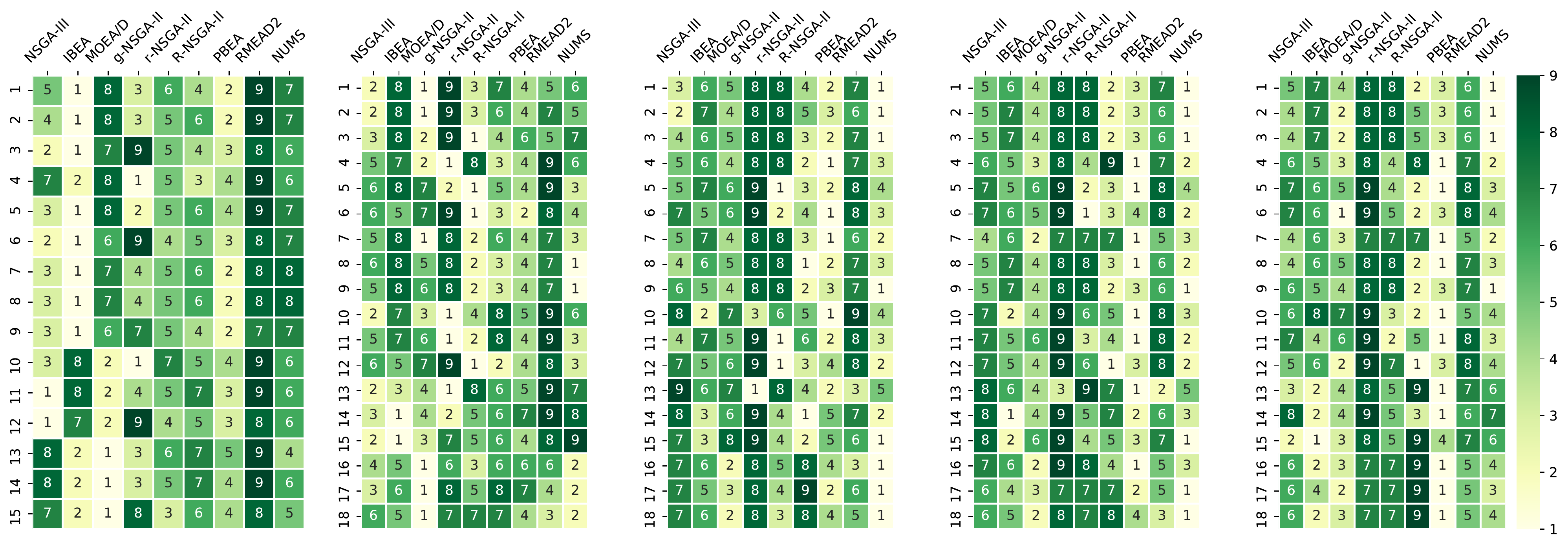}
    \caption{Heat maps of the ranks of R-HV values obtained by different algorithms for the \lq extreme\rq\ reference point settings. The subplots, from left to right, are ZDT problems and DTLZ problems with 3 to 10 objectives, respectively.}
    \label{fig:RHV_EXT}
\end{figure*}

Let us start our discussion from the 2-objective cases which are more intuitively visible. From the population plots shown in Figures 2 to 11 in the supplementary document, we find that NSGA-III and MOEA/D do not have any difficulty to approximate the whole PF; whilst IBEA has some trouble on ZDT2, ZDT3 and ZDT4. In particular, solutions obtained by IBEA can only cover a partial region of the PF, which may be outside of the expected ROI. These observations are contingent upon the promising R-IGD and R-HV metric values obtained by these non-preference based EMO algorithms as the heat maps of ranks shown in Figs.~\ref{fig:RIGD_BAL} to \ref{fig:RHV_EXT}. In particular, each cell of these figures represents the rank of the corresponding algorithm on a test instance. Whilst the numbers along the vertical axis are the shortcuts for the test instances where 3 rows form a group representing the $\mathbf{z}^r_\texttt{F}$, $\mathbf{z}^r_\texttt{I}$ and $\mathbf{z}^r_\texttt{P}$ settings on a particular test problem respectively. Although the preference-based EMO algorithms are designed to approximate solutions lying on the vicinity with respect to $\mathbf{z}^r$, as shown in Figures 2 to 11 in the supplementary document, their approximated solutions have shown some offsets with respect to the ROI. Notice that these observations depend on the PF shapes. These observations explain the large variance of the R-IGD and R-HV values obtained by those preference-based EMO algorithms. On the other hand, if we consider the $\mathbb{E}(P)$ metric, as shown in Figs.~\ref{fig:EP_BAL} and \ref{fig:EP_EXT}, the performance of the non-preference based EMO algorithms are not as competitive as those on the R-IGD and R-HV. This observation can be explained as the guidance provided by $\mathbf{z}^r$. Therefore, some preference-based EMO algorithms can have a better approximation to the DM most preferred solution, i.e. the one closest to $\mathbf{z}^r$.

Let us move to the higher-dimensional cases. As shown in Figures 12 to 59 along with the performance metrics shown in Tables 5 to 34 in the supplementary document, the performance of non-preference based EMO algorithms in the 3-objective case is not as competitive as the 2-objective scenario. In particular, IBEA can only find some special solutions (e.g. extreme points or boundary solutions) in most cases. In contrast, the superiority of some preference-based EMO algorithms becomes more evident with the increase of dimensionality. This can be explained as the expansion of the size of the PF with the dimensionality. In this case, solutions obtained by the non-preference-based EMO algorithms are sparsely distributed in a high-dimensional space. In other words, the chance for covering, by using a limited number of points, the expected ROI gradually decreases with the dimensionality. Moreover, as reported in many recent research (e.g.~\cite{IshibuchiAN15}), solving a many-objective optimisation problem itself is very challenging.

\begin{tcolorbox}[breakable, title after break=, height fixed for = none, colback = blue!20!white, boxrule = 0pt, sharpish corners, top = 0pt, bottom = 0pt, left = 2pt, right = 2pt]
    \underline{Answers to \textit{RQ}1}: Incorporating preference information into an EMO algorithm does not always lead to a better approximation to the ROI comparing to those traditional EMO algorithms, especially when the number of objectives is small. However, with the increase of dimensionality, incorporating preference information into the search process gradually becomes important. Due to the guidance provided by the DM supplied reference point(s), a preference-based EMO algorithm can have a better selection pressure towards the ROI. Furthermore, this is also beneficial to approximate the solution(s) most preferred by the DM, i.e. the one(s) closest to the DM supplied reference point.
\end{tcolorbox}

\subsection{Performance Comparisons of Different Preference-based EMO Algorithms}
\label{sec:response_RQ2}

As discussed in~\pref{sec:response_RQ1}, we appreciate the effectiveness of incorporating the DM's preference information for approximating the ROI. However, according to the results, we notice that not all preference-based EMO algorithms are able to have a desirable approximation to the ROI. In particular, some algorithms, where the DM's preference information is not appropriately utilised, were outperformed by non-preference-based EMO algorithms. 

\begin{figure}[htbp]
    \centering
    \includegraphics[width=.8\linewidth]{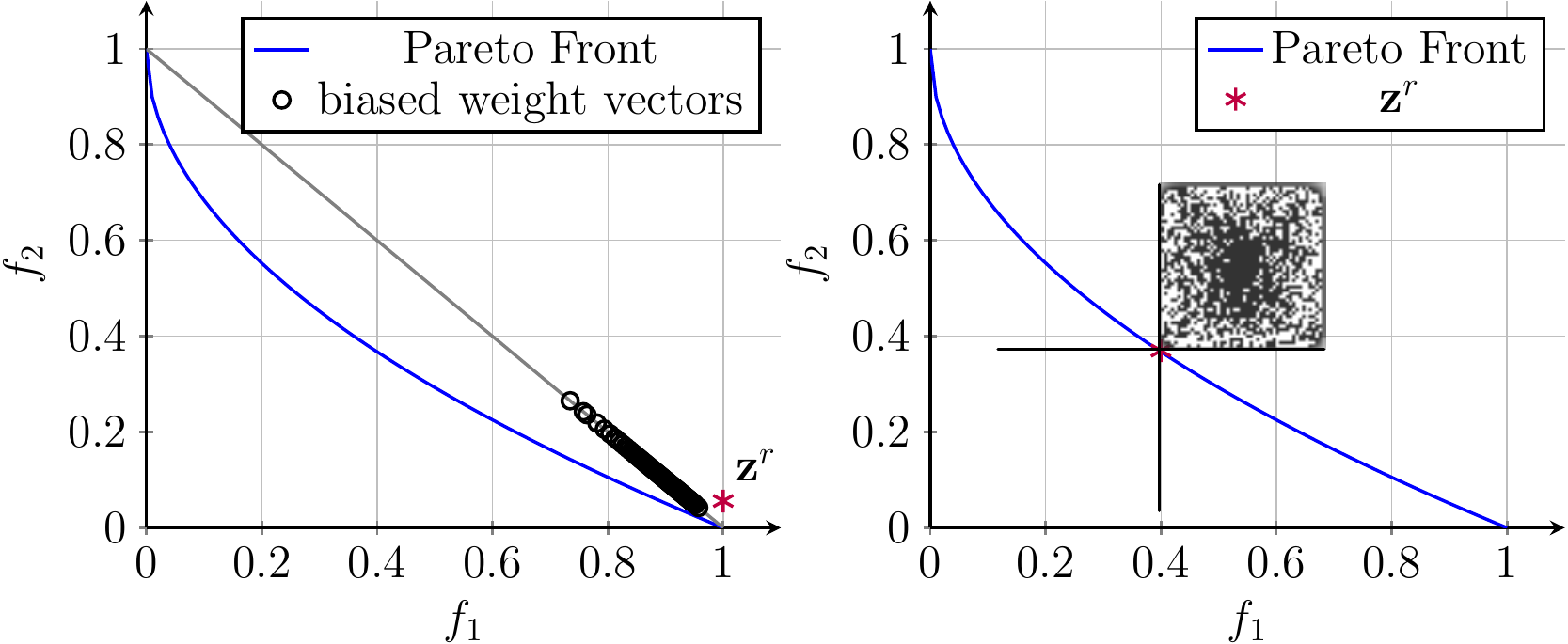}
    \caption{Illustrative examples of NUMS and g-dominance.}
    \label{fig:examples}
\end{figure}

Let us first look into MOEA/D-NUMS whose performance is constantly superior across all test cases, especially on the $\mathbb{E}(P)$ metric. Because the NUMS considers the projection of the DM supplied reference point on the simplex as one of the final biased weight vectors, it has a larger chance to find the solution most preferred by the DM. Furthermore, due to its theoretical guarantee, the NUMS is always able to generate a set of biased weight vectors with a given extent. This property will not be influenced by the problem dimensionality. However, we notice that MOEA/D-NUMS can hardly find the extreme point(s) on the PF when the DM supplied reference point is placed on one side the PF (as shown in Figures 2 to 11 in the supplementary document). As the example shown in the left panel of~\pref{fig:examples}, all reference points will be shifted towards the projection of the DM supplied reference point along the simplex, so that the extreme point is missed. In addition, there  is a tail extending towards the other end of the PF due to the non-uniform mapping. In contrast, the performance of RMEAD2, the other decomposition-based algorithm, is almost one of the worst among six preference-based EMO algorithms. Note that the weight vectors used in RMEAD2 gradually evolve towards the ROI with the population. Because the population evolution has some oscillations, it can be misleading to the adjustment of weight vectors. Moreover, as discussed in~\cite{GiagkiozisPF13}, frequently adjusting the distribution of weight vectors on-the-fly is negative to the search process of a decomposition-based EMO algorithm.

As for the three $\ast$-NSGA-II algorithms, their performance is similar in the 2-objective cases. Specifically, the selection pressure of g-NSGA-II comes from the box region specified by the DM supplied reference point. This is easy to implement in the 2-objective scenario. However, the effective area specified by the box region significantly decreases with the increase of dimensionality. It can hardly provide sufficient selection pressure towards the ROI when the number of objectives is larger than two. This effect is similar to the original Pareto dominance, and it explains the inferior performance of g-NSGA-II in the 3- to 10-objective cases. Furthermore, it is worth noting that g-NSGA-II becomes ineffective when the reference point is set exactly on the PF. As shown in the right panel of~\pref{fig:examples}, no solution will survive at the end except the DM supplied reference point. This effect has also been reflected by its poor performance when using a $\mathbf{z}^r_{\texttt{P}}$ setting. The selection mechanisms of R-NSGA-II and r-NSGA-II are similar. Their major difference is: R-NSGA-II directly uses the Euclidean distance towards the DM supplied reference point to guide the selection; whilst the r-NSGA-II has a parameter $\delta$ to control the comparability of two disparate non-dominated solutions. As a result, R-NSGA-II has shown more robust performance compared to r-NSGA-II. In addition, as Figures 12 to 59 shown in the supplementary document, we notice that the extent of the approximated ROI obtained by either R-NSGA-II or r-NSGA-II is ad-hoc. There is no thumb-rule to set an appropriate parameter to control this extent.

Different from the other preference-based EMO algorithms, which depend either on the Euclidean distance towards the DM supplied reference point or a set of biased weight vectors, PBEA uses a Pareto-compliant indicator to assign fitness to each solution. In particular, the Pareto-compliant property is guaranteed by both the binary indicator and the ASF. According to the results shown in Figs.~\ref{fig:EP_BAL} to~\ref{fig:RHV_EXT}, we find that PBEA is very competitive across all scenarios, especially when the number of objectives becomes large, e.g. $m=10$. Furthermore, according to the plots of population distribution shown in the supplementary document, we also notice that the extent of the approximated ROI found by PBEA is very narrow. In fact, the width of the approximated ROI is controlled by the specificity parameter $\delta$. But no thumb rule is available to set an appropriate $\delta$ for the desirable extent of the ROI. This is similar to R-NGSA-II and r-NSGA-II. %Furthermore, we also notice that the obtained solutions on the problem(s) with a convex PF shape (e.g. ZDT1) have a bias away from the DM supplied reference point.

\begin{figure}[htbp]
    \centering
    \includegraphics[width=.6\linewidth]{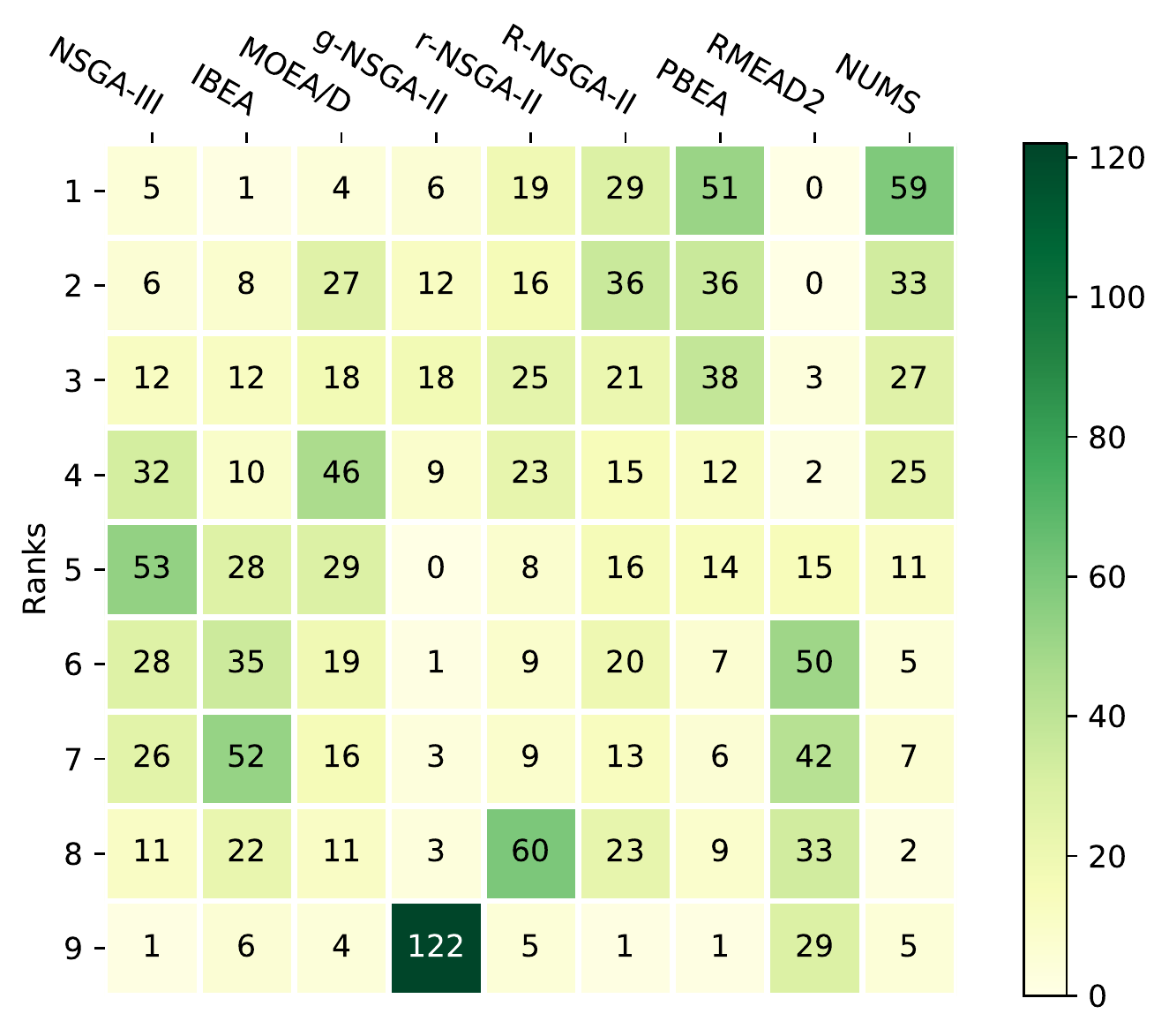}
    \caption{Heat maps of the number of times an algorithm has been ranked as a particular position in performance comparison.}
    \label{fig:whole_rank}
\end{figure}

To have a better overall picture of the performance of different algorithms, we summarise the ranking results across all test instances and plot them as the heat map shown in~\pref{fig:whole_rank}. In particular, each cell of this heat map represents the number of times the corresponding algorithm has been ranked as a particular position in the performance comparison. For example, the cell $(1, 9) = 59$ indicates that MOEA/D-NUMS has been ranked as the first place for 59 times. From these comparison results, we find that some preference-based EMO algorithms (i.e. R-NSGA-II, PBEA and MOEA/D-NUMS) have shown superior performance than those non-preference-based counterparts. This observation also supports the findings in the response to RQ1. On the other hand, we also find that some preference-based EMO algorithms (i.e. g-NSGA-II, r-NSGA-II and RMEAD2) are ranked as the worst algorithms across all test instances.

\begin{tcolorbox}[breakable, title after break=, height fixed for = none, colback = blue!20!white, boxrule = 0pt, sharpish corners, top = 0pt, bottom = 0pt, left = 2pt, right = 2pt]
    \underline{Answers to \textit{RQ}2}: From our experiments, we find that R-NSGA-II,  PBEA and MOEA/D-NUMS are the most competitive preference-based EMO algorithms for approximating various SOI. Notice that these three algorithms are based on different EMO frameworks. Thus, we conclude that dominance-, indicator- and decomposition-based EMO frameworks are all useful for approximating the SOI, given the preference information supplied by the DM is well utilised. In particular, transforming preference information into a distance metric (e.g. Euclidean distance or Tchebycheff distance) is a reliable way to guide the search towards the SOI. Otherwise, considering the DM's preference information can even lead to a negative effect to the search process as analysed in~\pref{sec:response_RQ2}. In particular, from our experiments, we can see that g-NSGA-II and RMEAD2 are even worse than those non-preference based EMO algorithms in most cases. Moreover, almost all algorithms, except g-NSGA-II, claimed that the approximated ROI is controllable by some specific parameter(s). However, only MOEA/D-NUMS provides a tangible way to control the size of ROI; whilst the others are all set in an ad-hoc manner.
\end{tcolorbox}

\vspace{-1em}

\subsection{Influence of the Location of Reference Points}
\label{sec:location}

%To this end, we choose to use reference points outside the box region specified by the ideal and nadir points.
In the previous experiments, we find that preference-based EMO algorithms can have a decent approximation to the ROI if the DM supplied preference information is used in an appropriate manner. A natural question arises: \textit{what happens if the DM supplies a \lq bad\rq\ preference information that does not represent her/his actual aspiration?} In many real-world scenarios, it is not rare that the DM has little knowledge about the underlying black-box system at the outset of the optimisation process. Therefore, it is not trivial to set an appropriate reference point that perfectly represents the DM's preference information. In this subsection, we will investigate the influence of the setting of reference point, i.e. its location, on the performance of preference-based EMO algorithms. For proof of concept purpose and to facilitate a better visual understanding, here we only conduct experiments on 2- and 3-objective cases whilst the conclusions are able to be generalised to problems with a larger number of objectives according to our preliminary experiments.

\begin{figure}[htbp]
\centering 
\includegraphics[width=1\linewidth]{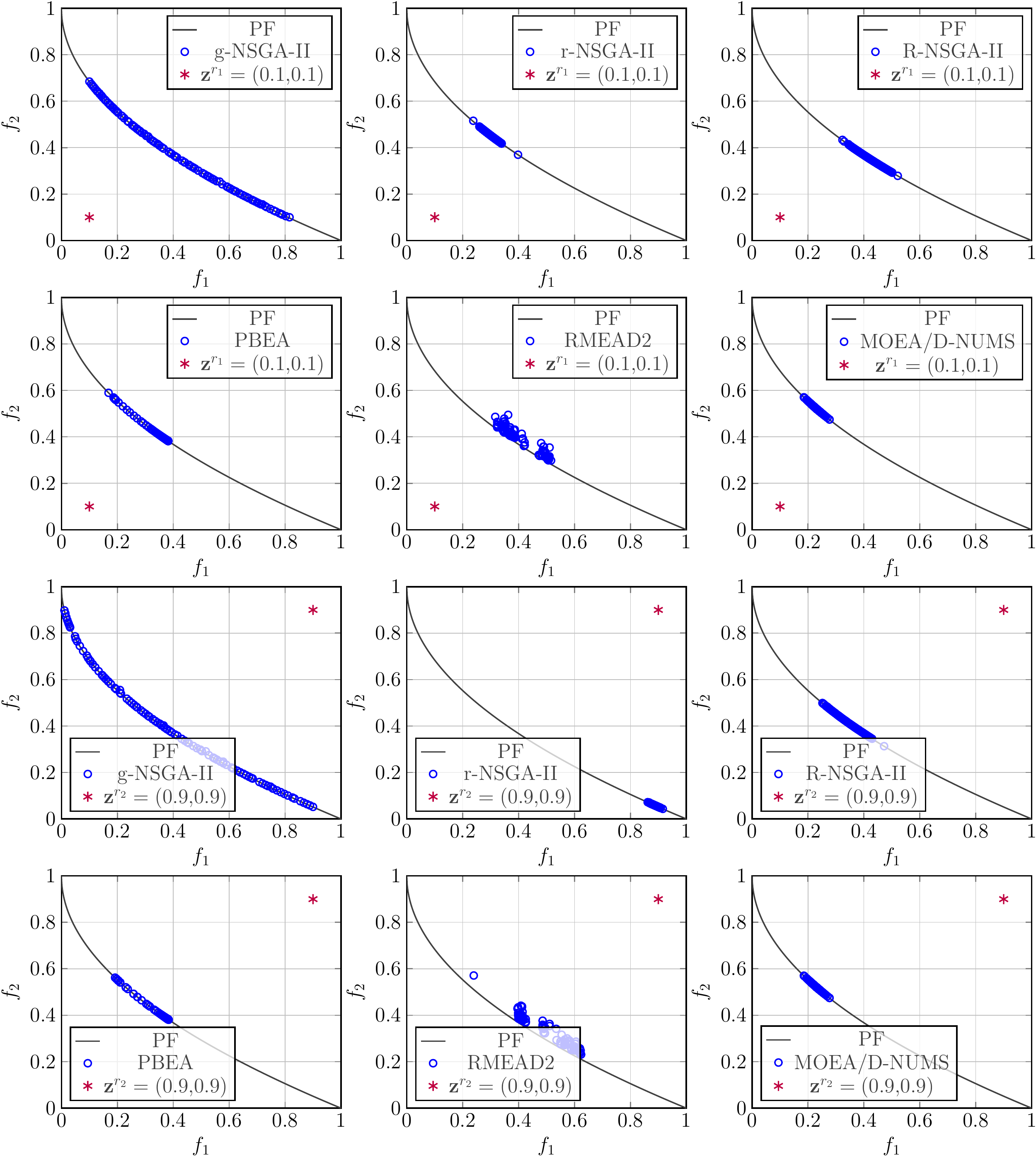}
    \caption{Solutions obtained by six preference-based EMO algorithms on the ZDT1 test problem when setting $\mathbf{z}^{r_1}=(0.1,0.1)^T$ and $\mathbf{z}^{r_2}=(0.9,0.9)^T$.}
\label{fig:bad_example1}
\end{figure}

Let us first look at two examples on ZDT1 where we consider two extreme reference point settings far away from the PF: $\mathbf{z}^{r_1}=(0.1, 0.1)^T$ in the infeasible region and $\mathbf{z}^{r_2}=(0.9,0.9)^T$ in the feasible region. \pref{fig:bad_example1} plots the solutions obtained by six preference-based EMO algorithms with the best R-IGD values. From these results, we find that R-NSGA-II, PBEA and RMEAD2 work as usual. In particular, RMEAD2 normally cannot find well converged solutions. On the other hand, although the solutions obtained by r-NSGA-II and MOEA/D-NUMS well converge to the PF, they all show certain mismatch with respect to the ROIs. As for g-NSGA-II, its solutions almost cover the entire PF. As discussed in \pref{sec:response_RQ2} and \pref{fig:examples}, the effective region of g-dominance is the box region covered by the DM supplied reference point. The farther the reference point away from the PF, the larger region covered by the reference point.

\begin{figure}[htbp]
\centering 
\includegraphics[width=1\linewidth]{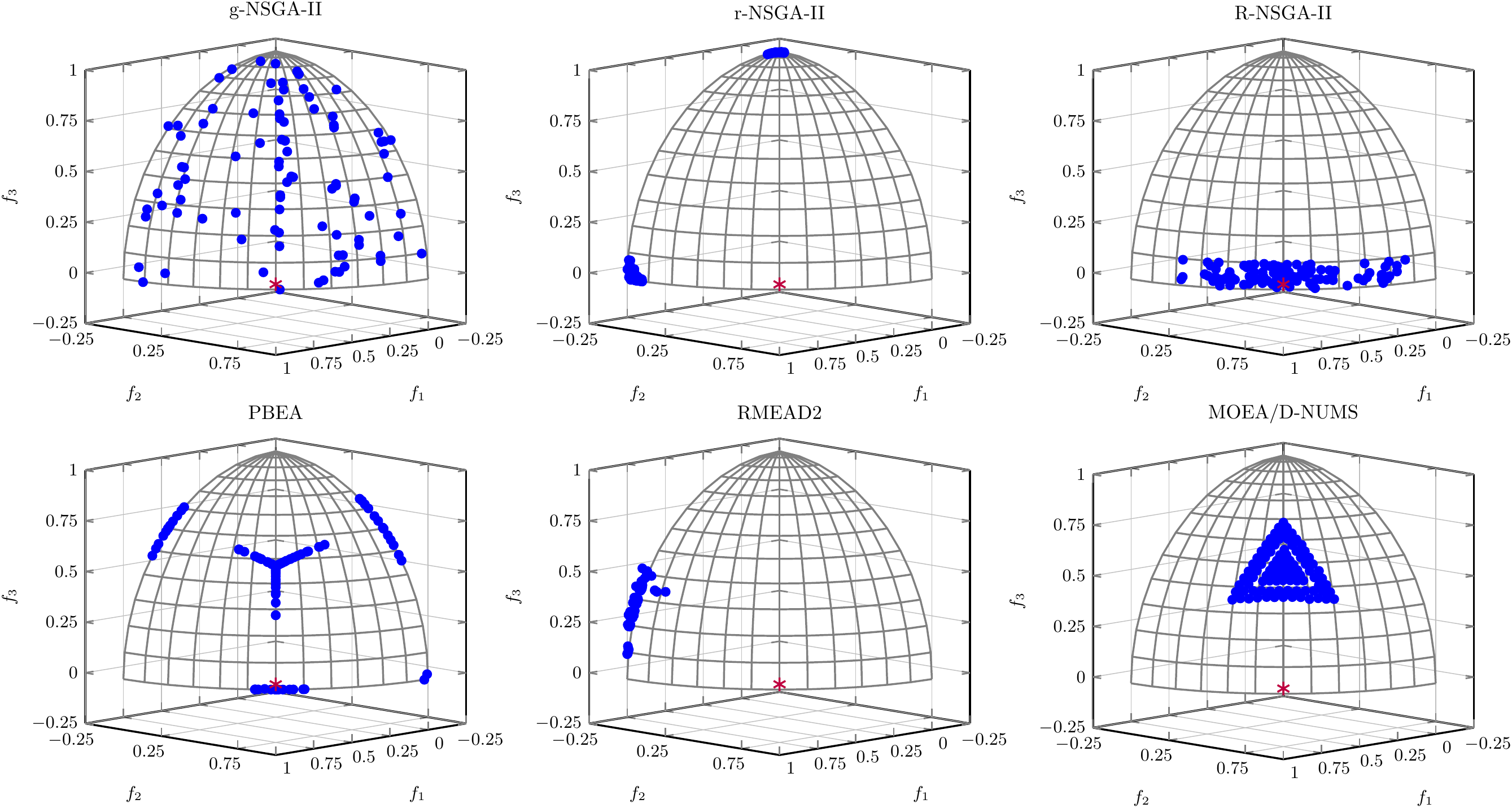}
\caption{Solutions obtained by six preference-based EMO algorithms on the DTLZ2 problem where $\mathbf{z}^{r_3}=(-0.2,-0.2,-0.2)^T$.}
\label{fig:bad_example3}
\end{figure}

Let us look at another example on DTLZ2 with three objectives. Here we set the reference point as $\mathbf{z}^{r_3}=(-0.2,-0.2,-0.2)^T$ which dominates the ideal point. In particular, one may argue that the DM will not set negative values as a reference point. In this case, we assume that such reference point setting represents that the DM expects for solutions having a as good objective value as possible at each objective. From the experimental results shown in~\pref{fig:bad_example3}, we can see that almost all algorithms, except MOEA/D-NUSM, have shown some unexpected behaviour. Specifically, g-NSGA-II and PBEA almost degenerate to their non-preference-based baseline EMO algorithm, i.e. NSGA-II and IBEA, as the obtained solutions tend to cover the entire PF. r-NSGA-II, R-NSGA-II and RMEAD2 end up with solutions lying on a boundary of the PF in an add-hoc manner.

\begin{tcolorbox}[breakable, title after break=, height fixed for = none, colback = blue!20!white, boxrule = 0pt, sharpish corners, top = 0pt, bottom = 0pt, left = 2pt, right = 2pt]
    \underline{Answers to \textit{RQ}3}: From our experiments, we find that a preference-based EMO algorithm may not work as expected given a \lq bad\rq\ reference point. In particular, a so called \lq bad\rq\ choice is typically a reference point way beyond the PF. In this case, the DM supplied reference point is either far away the optima (s)he actually expects or too utopian to approach. In real-world black-box optimisation scenarios, it is not rare that the DM struggles to set a reasonably good reference point given her/his little knowledge about the underlying problem. This becomes even severer when having a large number of objectives.
\end{tcolorbox}

\vspace{-1em}

\subsection{Incorporating User Preference in an Interactive Manner}
\label{sec:interactive}

The previous experiments are conducted under the \textit{a priori} elicitation manner. As discussed in~\pref{sec:interactive_literature}, using a reference point to represent the DM's preference information can be directly used in an \textit{interactive} manner. Different from many studies on preference-based EMO in the literature (e.g. \cite{DebSBC06,SaidBG10,LuqueSHCC09,ThieleMKL09,MohammadiOLD14,LiCMY18}), which are mainly tested on benchmark problems, this paper considers testing the effectiveness of interactive EMO on stock market portfolio optimisation under a real-world setting. In particular, we collect the stock market data of 58 listed companies from Shenzhen Stock Exchange A Share since 1990. Two popular portfolio optimisation models are considered in this paper.

The first one is the Mean-Variance-Skewness (MVS) model proposed by Konno and Suzuki~\cite{KonnoS95}. Specifically, given a portfolio of financial assets $\mathbf{P}=(\rho_1,\cdots,\rho_n)^T$ where $\rho_i$ indicates the percentage of the wealth invested in the $i$-th asset and $\sum_{i=1}^n\rho_i=1$, the return of $\mathbf{P}$ is calculated as:
\begin{equation}
    \psi[\mathbf{P}]=\sum_{i=1}^n\rho_ir_i,
\end{equation}
where $r_i$ is the rate of return of $\rho_i$, $i\in\{1,\cdots,n\}$. The MVS model is formulated as:
\begin{equation}
    \left \{
        \begin{array}{cc}
            \mathrm{maximise} &\mathbb{E}[\psi(\mathbf{P})]=\sum_{i=1}^n\rho_i\mathbb{E}[r_i] \\
            \mathrm{minimise} & \mathbb{V}[\psi(\mathbf{P})]=\mathbb{E}[(\psi(\mathbf{P})-\mathbb{E}[\psi(\mathbf{P})])^2] \\
            \mathrm{maximise} &\mathbb{S}[\psi(\mathbf{P})]=\mathbb{E}[(\psi(\mathbf{P})-\mathbb{E}[\psi(\mathbf{P})])^3]
        \end{array}
        \right.,
        \label{MVS}
\end{equation}

In the experiments, only R-NSGA-II, PBEA and MOEA/D-NUMS are chosen as the preference-based EMO algorithms given their superior performance reported in~\pref{sec:response_RQ2}. The population size is set to 91 for MOEA/D and MOEA/D-NUMS and 92 for the others; whilst the maximum number of function evaluations is set to 5,520, i.e. approximately 60 generations. The DM is assumed to have three chances to elicit her/his preference information. Because the PF is unknown, only the R-HV metric is chosen in the performance assessment.

\begin{figure}[htbp]
    \centering
    \includegraphics[width=\linewidth]{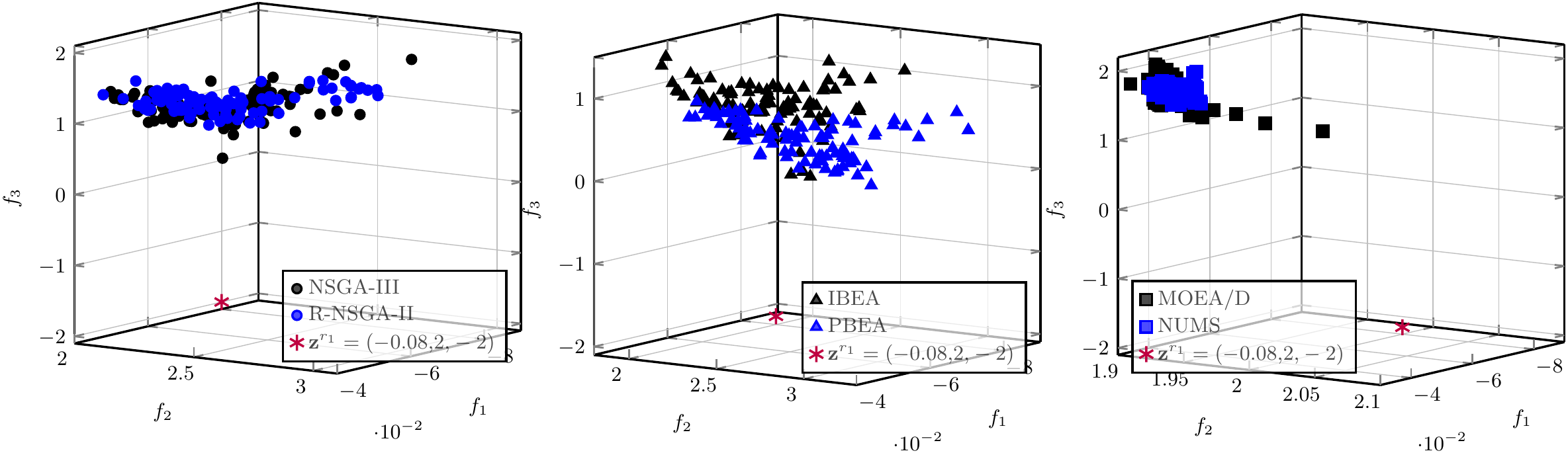} 
    \caption{Solutions obtained on the 3-objective portfolio optimisation problem in the first interaction, where $\mathbf{z}^{r_1}=(-0.08,2,-2)^T$.} 
    \label{fig:IEMO_M3_1}
\end{figure}

\begin{figure}[htbp]
    \centering
    \includegraphics[width=\linewidth]{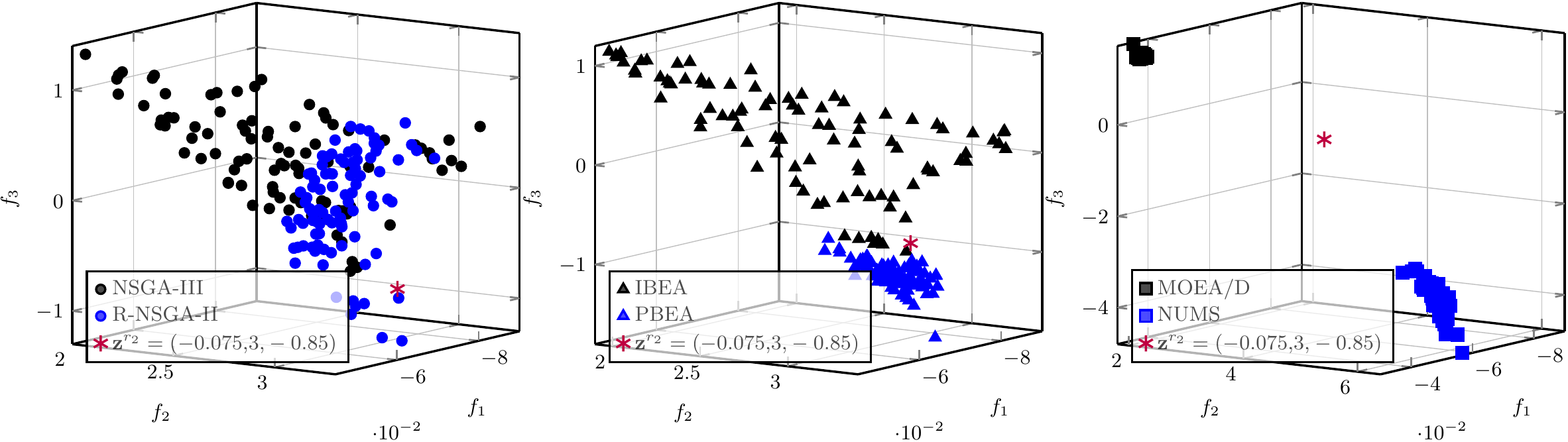} 
    \caption{Solutions obtained on the 3-objective portfolio optimisation problem in the second interaction, where $\mathbf{z}^{r_2}=(-0.75,3,-0.85)^T$.} 
    \label{fig:IEMO_M3_2}
\end{figure}

\begin{figure}[htbp]
    \centering
    \includegraphics[width=\linewidth]{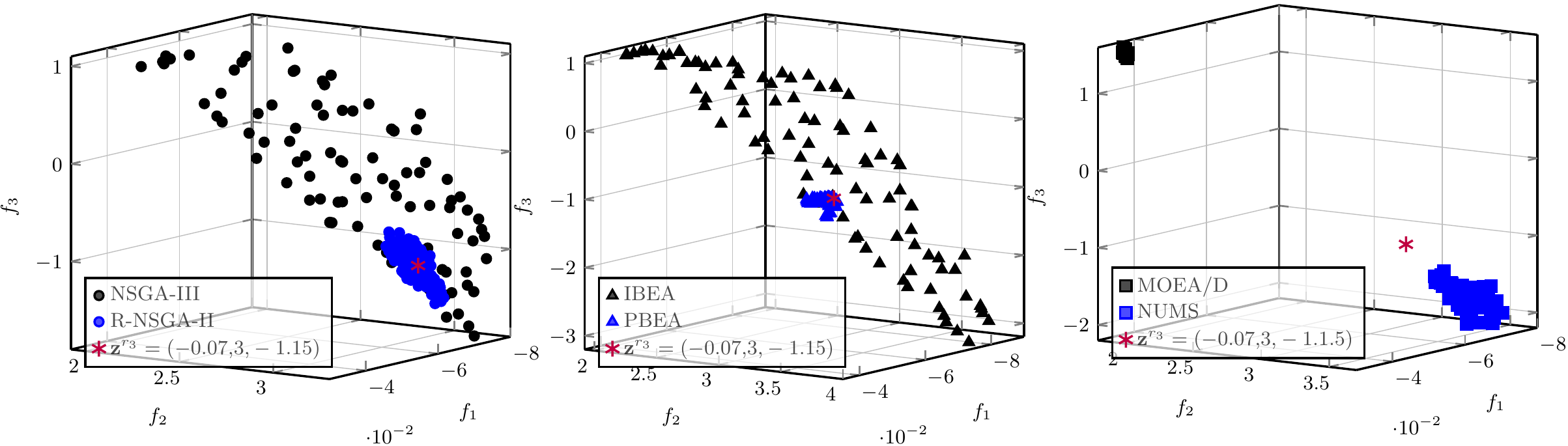} 
    \caption{Solutions obtained on the 3-objective portfolio optimisation problem in the third interaction, where $\mathbf{z}^{r_3}=(-0.07,3,-1.15)^T$.} 
    \label{fig:IEMO_M3_3}
\end{figure}

\begin{figure}[htbp]
    \centering 
    \includegraphics[width=.5\linewidth]{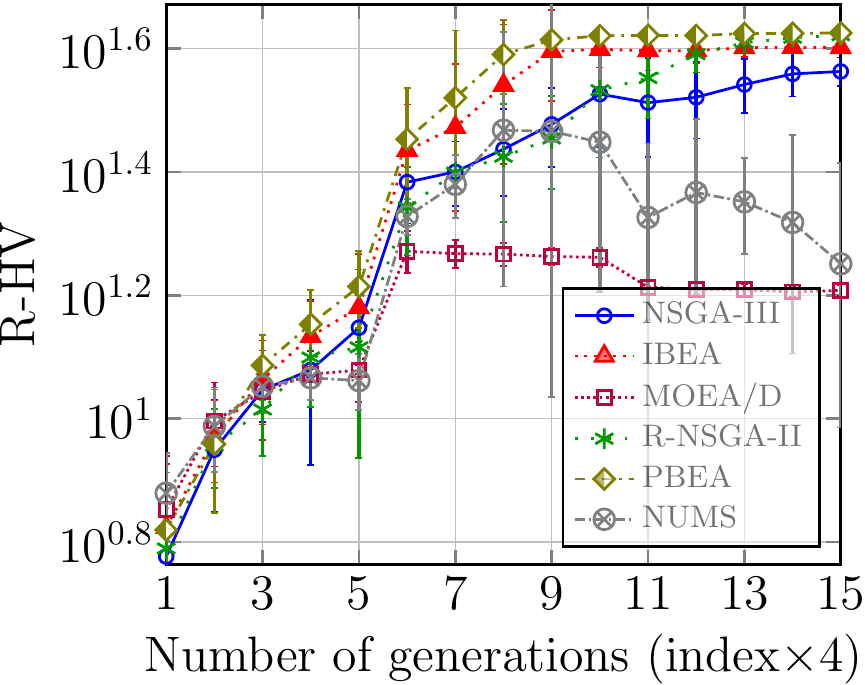} 
    \caption{Trajectories of R-HV values versus the number of generations on the 3-objective portfolio optimisation problem.} 
    \label{fig:traj_M3}
\end{figure}

Solutions obtained by different algorithms after three preference elicitations are presented in Figs.~\ref{fig:IEMO_M3_1} to~\ref{fig:IEMO_M3_3}. More specifically, in the first preference elicitation, we assume that the DM is rather greedy. (S)he sets $\mathbf{z}^{r_1}=(-0.75,3,-0.85)^T$ where each objective is as utopia as possible. As shown in~\pref{fig:IEMO_M3_1}, solutions obtained by the preference-based EMO algorithms are similar to their non-preference-based counterparts. It is interesting to note that the performance of MOEA/D is better than MOEA/D-NUMS according to the R-HV trajectories shown in~\pref{fig:traj_M3}. Furthermore, it is clear that almost all solutions are dominated by $\mathbf{z}^{r_1}$. This suggests that $\mathbf{z}^{r_1}$ is too utopia to achieve.

In the second preference elicitation, the DM made some modifications on some objectives and set $\mathbf{z}^{r_2}=(-0.75,3,-0.85)^T$. As shown in~\pref{fig:IEMO_M3_2}, solutions found by R-NSGA-II and PBEA have a much better approximation to $\mathbf{z}^{r_2}$ this time whilst solutions obtained NSGA-III and IBEA do not change significantly. In addition, as shown in~\pref{fig:traj_M3}, the trajectories of R-HV values have a significant surge after the elicitation of $\mathbf{z}^{r_2}$. This is partially caused by using a more reasonable reference point to guide the preference-based EMO algorithms and also in the performance evaluation.

Moreover, we also notice that $\mathbf{z}^{r_2}$ is dominated by some solutions obtained by R-NSGA-II and PBEA, this suggests that some objectives deserve better expectation. Bearing this consideration in mind, the DM fine-tunes the reference point and set $\mathbf{z}^{r_3}=(-0.07,3,-1.15)^T$ in the last preference elicitation. As shown in~\pref{fig:IEMO_M3_3}, solutions obtained by R-NSGA-II and PBEA have a decent approximation around $\mathbf{z}^{r_3}$. This is also reflected by their best R-HV values. In contrast, solutions found by NSGA-III and IBEA do not show significant difference with respect to the second preference elicitation. This suggests that they almost converge. Moreover, since $\mathbf{z}^{r_3}$ is almost on the PF manifold and the solutions obtained by NSGA-III and IBEA well approximate the whole PF, their R-HV values are also competitive. However, we notice that the performance of MOEA/D and MOEA/D-NUMS are even worse. This might be caused by the largely disparate scales of different objectives which make the simplex assumption of decomposition-based EMO algorithm fail to meet the actual shape of the PF.

In addition to the three objectives considered in the MVS model, investors may also consider the robustness and the portfolio return as additional objectives in their portfolio investments. As for the prior objective, we apply the kurtosis model proposed by Lai et al.~\cite{LaiYW06} to evaluate the probability of extreme events. In particular, the larger the kurtosis is, the higher probability the extreme events occur. In other words, the corresponding portfolio investment is less robust. Specifically, the kurtosis can be calculated as:
\begin{equation}
    \mathbb{K}[\psi(\mathbf{P})]=\mathbb{E}[(\psi(\mathbf{P})-\mathbb{E}(\mathbf{P}))^4],
\end{equation}
As for the portfolio return, it can be evaluated as the turnover of stock investments. In particular, a high turnover ratio indicates an active state of the underlying stock investments. Specifically, the turnover of a portfolio of financial assets $\mathbf{P}$ is calculated as:
\begin{equation}
    \phi(\mathbf{P})=\sum_{i=1}^n\rho_it_i,
\end{equation}
where $t_i$ represents the turnover of each financial asset $\rho_i$, $i\in\{1,\cdots,n\}$. The expected turnover is calculated as:
\begin{equation}
    \mathbb{E}(\phi(\mathbf{P}))=\sum_{i=1}^n\rho_i\mathbb{E}[t_i],
\end{equation}
In summary, the Mean-Variance-Skewness-Kurtosis-Turnover (MVSKT) model, which constitutes a five-objective portfolio optimisation problem, is formulated as:
\begin{equation}
    \left \{
        \begin{array}{cc}
            \mathrm{maximise} &\mathbb{E}[\psi(\mathbf{P})] \\
            \mathrm{minimise} & \mathbb{V}[\psi(\mathbf{P})] \\
            \mathrm{maximise} &\mathbb{S}[\psi(\mathbf{P})] \\
            \mathrm{minimise} &\mathbb{K}[\psi(\mathbf{P})] \\
            \mathrm{maximise} &\mathbb{E}[\phi(\mathbf{P})] \\
        \end{array}
        \right.,
        \label{MVSKT}
\end{equation}

\begin{figure}[htbp]
    \centering 
    \includegraphics[width=.9\linewidth]{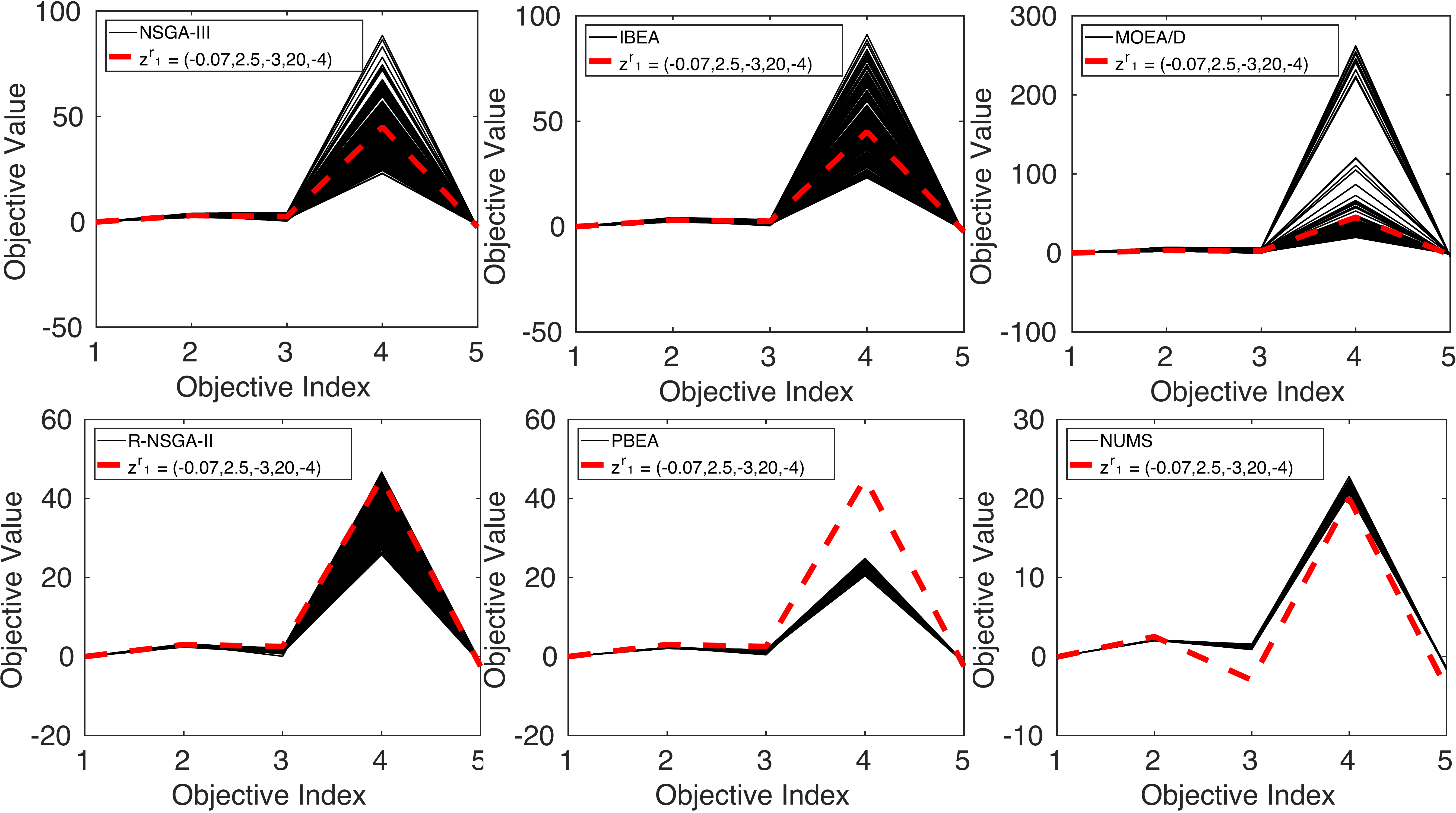} 
    \caption{Solutions obtained on 5-objective portfolio optimisation problem in the first interaction, where $\mathbf{z}^{r_1}=(-0.07,2.5,-3,20,-4)^T$.} 
    \label{fig:IEMO_M5_1}
\end{figure}

\begin{figure}[htbp]
    \centering 
    \includegraphics[width=.9\linewidth]{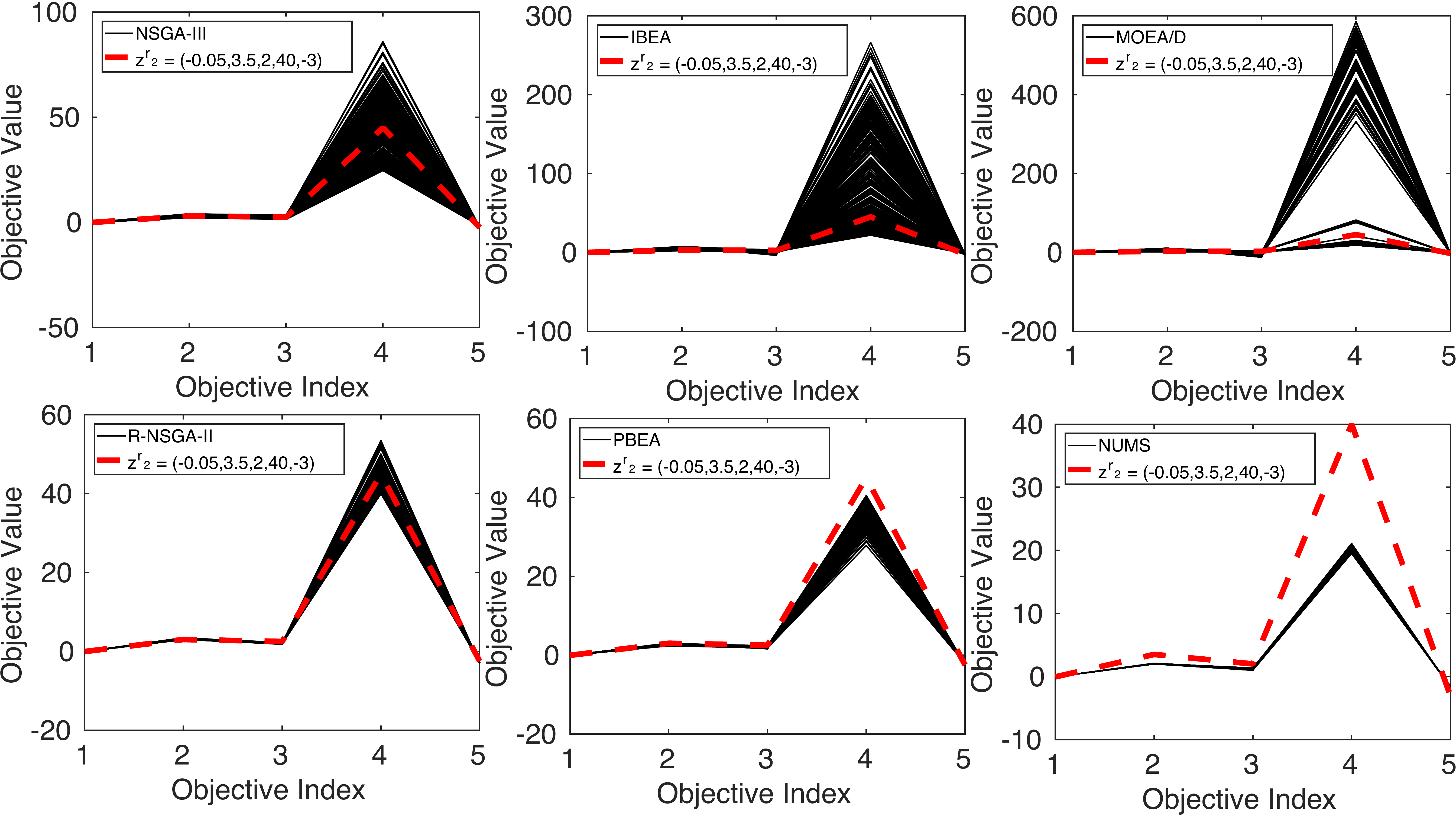} 
    \caption{Solutions obtained on 5-objective portfolio optimisation problem in the second interaction, where $\mathbf{z}^{r_2}=(-0.05,3.5,2,40,-3)^T$.} 
    \label{fig:IEMO_M5_2}
\end{figure}

\begin{figure}[htbp] 
    \centering 
    \includegraphics[width=.9\linewidth]{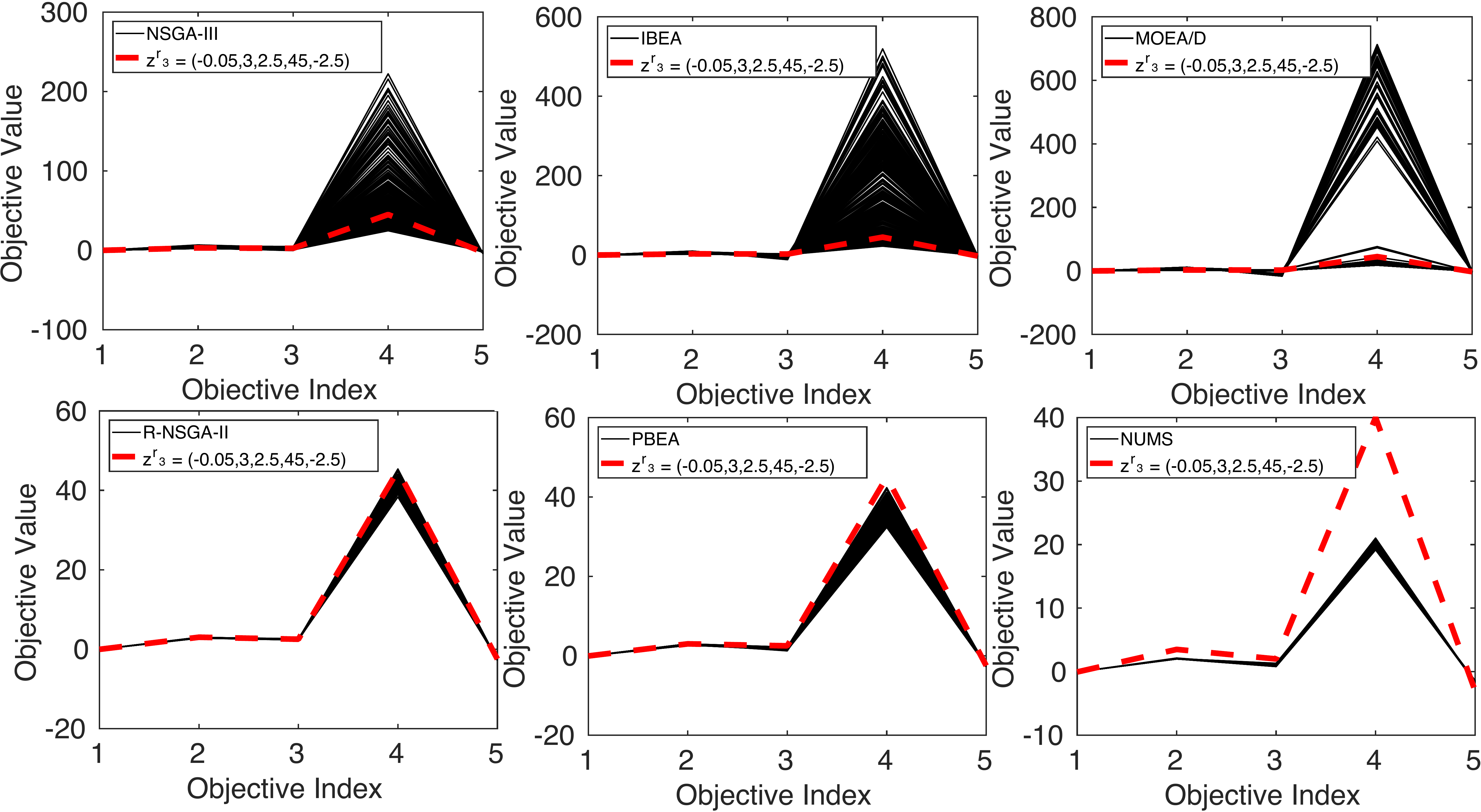} 
    \caption{Solutions obtained on 5-objective portfolio optimisation problem in the third interaction, where $\mathbf{z}^{r_3}=(-0.05, 3,2.5,45,-2.5)^T$.} 
    \label{fig:IEMO_M5_3}
\end{figure}

In the experiments, almost all settings are the same as the 3-objective case except the population size and the number of function evaluations. In particular, the population size is set to 210 for MOEA/D and MOEA/D-NUMS and 212 for the others; whilst the maximum number of function evaluations is set to 12,720, i.e. approximately 60 generations in total. Figs.~\ref{fig:IEMO_M5_1} to \ref{fig:IEMO_M5_3} plot the population distributions of solutions obtained by different algorithms after three preference elicitations.

\begin{figure}[htbp]
  	\centering 
	\includegraphics[width=.5\linewidth]{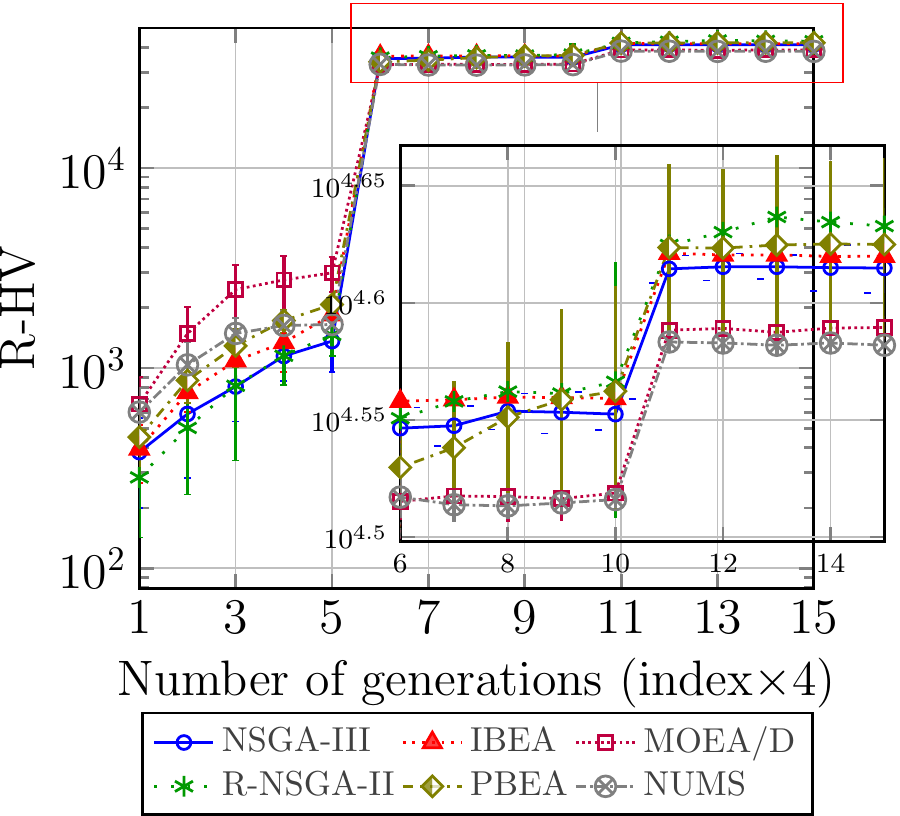} 
  	\caption{Trajectories of R-HV values versus the number of generations on the 5-objective portfolio optimisation problem.} 
	\label{fig:traj_M5}
\end{figure}

Similar to the three-objective case, we assumed that the DM specifies a reference point which has an as utopia value as possible at each objective. From~\pref{fig:IEMO_M5_1} and~\pref{fig:traj_M5}, we find that three preference-based EMO algorithms have shown similar performance in terms of population distribution and R-HV values. In particular, MOEA/D is the best algorithm under such preference setting.

In the second preference elicitation, the DM modified the aspiration at each objective, especially on the skewness and kurtosis. As shown in~\pref{fig:traj_M5}, all R-HV trajectories have experienced a significant surge after the second preference elicitation. This is similar to the observation in~\pref{fig:traj_M3}. It is also interesting to note that although the R-HV values of MOEA/D and MOEA/D-NUMS have been improved, their obtained solutions are not as satisfactory as the other peers. Especially for MOEA/D-NUMS, its obtained solutions do not have significant difference comparing to the first preference elicitation.

As shown in~\pref{fig:traj_M5}, the R-HV values were improved in the last preference elicitation. As discussed before, this is partially because the DM supplied reference point becomes more reasonable. As shown in~\pref{fig:IEMO_M5_3}, solutions found by R-NSGA-II and PBEA are close to $\mathbf{z}^{r_3}$. In contrast, solutions found by MOEA/D and MOEA/D-NUMS do not show significant difference with respect to the change of reference point.

%, i.e. at the outset and every 20 generations respectively afterwards
%由Fig？的trajectory图可以看出，在每次输入DM的preference information之后，preference-based EMO都能比较好的逼近到尽可能满足DM preference的区域的解，而对于那些non-preference based EMO，他们只是尽力不断地去逼近整个PF。必然不会对DM的preference进行响应，亦不会随之调整自身的搜索行为。
\begin{tcolorbox}[breakable, title after break=, height fixed for = none, colback = blue!20!white, boxrule = 0pt, sharpish corners, top = 0pt, bottom = 0pt, left = 2pt, right = 2pt]
    \underline{Answers to \textit{RQ}4}: From our experiments, we find that a preference-based EMO algorithm is able to respond to the change of the DM's preference information in an interactive manner. As discussed in~\pref{sec:location}, eliciting appropriate preference information is far from trivial, especially under a black-box setting. In other words, the DM may easily elicit an unrealistically utopian aspiration for each objective function at the outset. An interactive elicitation manner provides the DM with an avenue to progressively understand the underlying problem and adjust her/his preference information within limited computational budgets.
\end{tcolorbox}

\subsection{Using Preference-based EMO Algorithms as a General-Purpose Optimiser}
\label{sec:massive_obj}

In our previous experiments, the preference-based EMO algorithms are studied in a conventional way, i.e. used to approximate a ROI, which is normally a partial region of the PF. On the other hand, we come up with another question: \textit{can we expect a preference-based EMO algorithm to be capable of approximating the whole PF if we set more than one ROI evenly spreading over the PF?} To address this question, we choose R-NSGA-II and PBEA as the representative preference-based EMO algorithms in our experiments to compare with three non-preference-based EMO algorithms, i.e. NSGA-III, IBEA and MOEA/D. In particular, we do not consider MOEA/D-NUMS because it becomes an ordinary MOEA/D when used to approximate the whole PF; whilst g-NSGA-II, r-NSGA-II and RMEAD2 are not considered given their poor performance reported in Sections~\ref{sec:response_RQ1} and~\ref{sec:response_RQ2}.

DTLZ1 to DTLZ4 are used as the benchmark problems where the number of objectives is set as $m\in\{3, 5, 8, 10, 15, 25, 50, 100\}$. In our experiment, we use a set of evenly distributed weight vectors, as used in the decomposition-based EMO methods, to represent the preference that cover the whole PF. As discussed in our recent study on massive objective optimisation~\cite{LiDAY17}, it is very challenging to set evenly distributed weight vectors when $m\geq 25$; whilst we use the weight vector generation method proposed therein~\cite{LiDAY17} to serve our purpose. Note that R-NSGA-II and PBEA are all able to handle more than one DM supplied reference point. Since we need to consider test problems with a massive number of objectives, the calculation of HV will be extremely time consuming even when using a Monte Carlo approximation~\cite{BringmannF13}. In this case, only the IGD metric is considered in our experiment to evaluate the performance. In particular, the settings of IGD calculation for problems with $m<25$ can be found in~\cite{LiDZK15}; otherwise we use the settings suggested in~\cite{LiDAY17}. Furthermore, the settings of the population size and the number of generations can be found in Section 1 the supplementary document.

\begin{figure}[htbp]
    \centering 
    \includegraphics[width=3.5in]{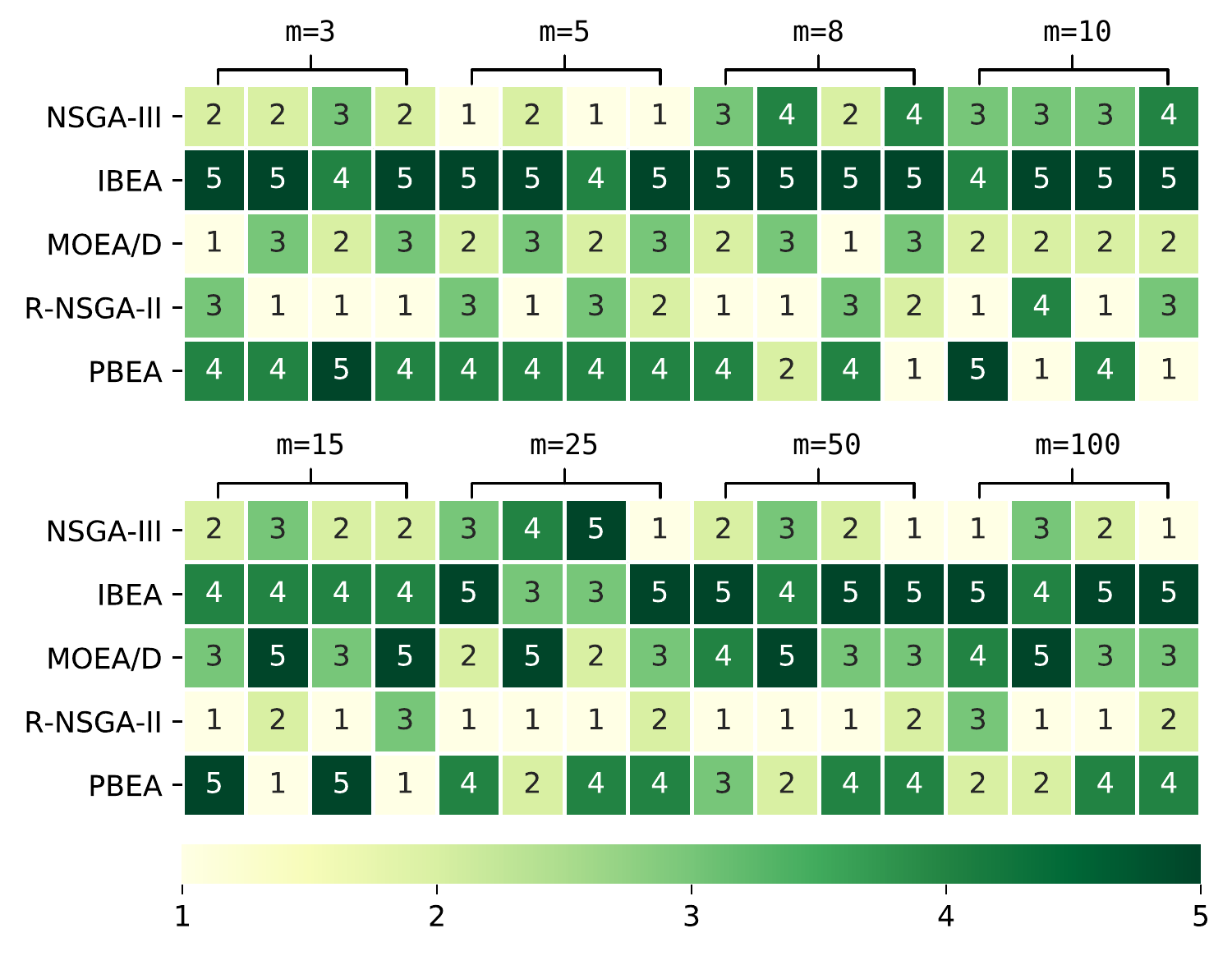} 
    \caption{Heat maps of the ranks of IGD values obtained by different algorithms on DTLZ1 to DTLZ4 problems with various numbers of objectives.} 
    \label{fig:massive_ranking}
\end{figure}

Similar to the previous subsections, the comparison results are presented as heat maps of ranks (shown in~\pref{fig:massive_ranking}) of IGD metric values obtained by different algorithms (detailed IGD metric values can be found in Table 36 in the supplementary document). From these results, we can see that NSGA-III, MOEA/D and R-NSGA-II are the most competitive algorithms; whilst IBEA is the worst one in most cases. Although the performance of PBEA is not promising when the number of objectives is relatively small, it gradually becomes more competitive with the increase of dimensionality. In principle, R-NSGA-II can be regarded as a decomposition-based algorithm when the DM supplied reference points are replaced by the weight vectors used in NSGA-III and MOEA/D. Their major difference lies in the way of how to evaluate the closeness of a solution to a weight vector. Specifically, it is evaluated as the perpendicular distance towards the reference line formed by the origin and a weight vector in NSGA-III. MOEA/D uses the Tchebycheff distance to evaluate the fitness of a solution. In R-NSGA-II, it uses the Euclidean distance between a solution and a weight vector as a major criterion in the environmental selection. It is interesting to note that R-NSGA-II has achieved the best IGD values in many cases whereas the distribution of its obtained solutions is not satisfactory as shown in Figs.~\ref{fig:massive_3D} and~\ref{fig:massive_100D} and Figures 60 to 91 in the supplementary document. This observation is not surprising as R-NSGA-II does not have a sophisticated diversity preservation mechanism and the using of more than one reference point may bring more uncertainty in selection process. We infer that its promising IGD values come from the better convergence of solutions towards the PF.

\begin{figure}[htbp]
    \centering 
    \includegraphics[width=.9\linewidth]{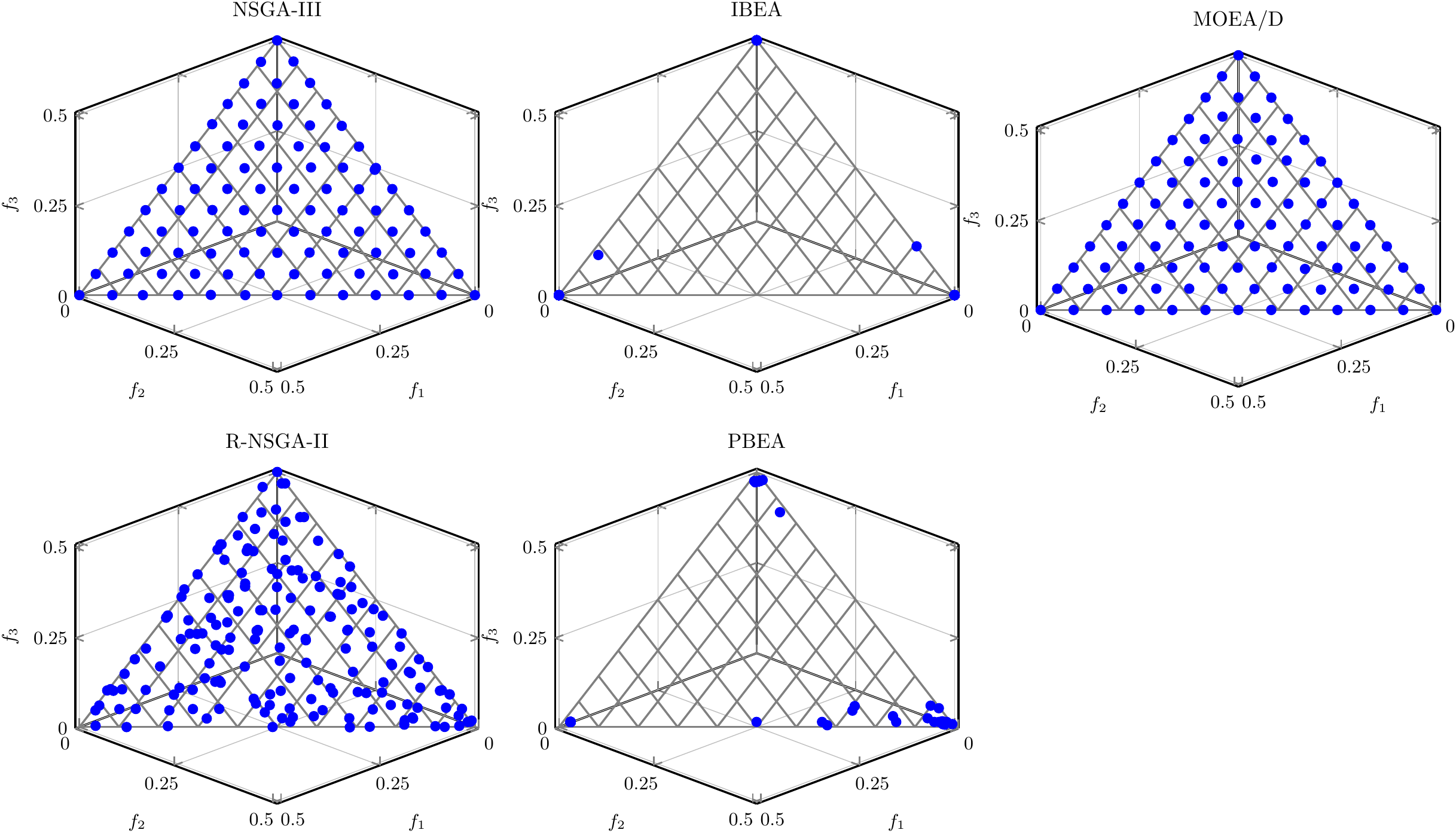} 
    \caption{Solutions obtained by different algorithms on the DTLZ1 test problem with 3 objectives.} 
    \label{fig:massive_3D}
\end{figure}

\begin{figure}[htbp]
    \centering 
    \includegraphics[width=.9\linewidth]{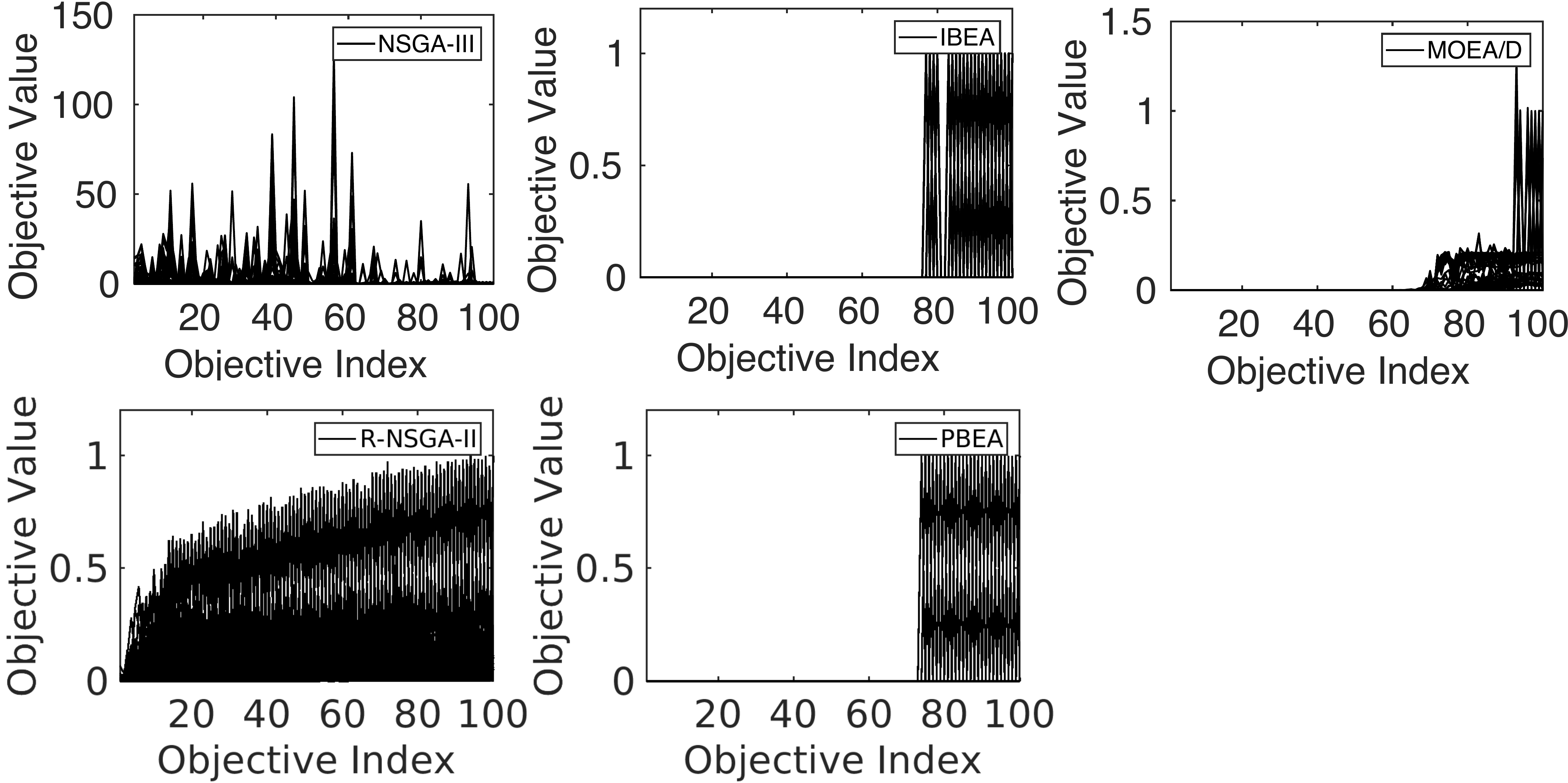} 
    \caption{Solutions obtained by different algorithms on the DTLZ3 test problem with 100 objectives.} 
    \label{fig:massive_100D}
\end{figure}

\begin{tcolorbox}[breakable, title after break=, height fixed for = none, colback = blue!20!white, boxrule = 0pt, sharpish corners, top = 0pt, bottom = 0pt, left = 2pt, right = 2pt]
    \underline{Answers to \textit{RQ}5}: From our experiments, we find that a preference-based EMO algorithm is able to approximate the whole PF, given that the DM supplied preference information is used in an appropriate manner. In particular, the direct Euclidean distance between the solution and the reference point, as in R-NSGA-II, is a surprisingly reliable metric to guide the optimisation process. By this means, preference elicitation become another decomposition method in EMO. Since there are more than one reference point, one key challenge is how to balance the search power across different reference points. Furthermore, it is also challenging to maintain the local diversity within a ROI specified by each reference point.
\end{tcolorbox}

%!TeX root=main.tex

\section{Conclusions and Future Works}
\label{sec:conclusions}

Finding trade-off solution(s) most satisfying the DM's preference information is the ultimate goal of multi-objective optimisation in practice. This paper first provides a pragmatic review on the current developments of preference-based EMO. In particular, the literature review was mainly conducted according to the elicitation manner, i.e. when to ask the DM to elicit her/his preference information. Afterwards, we conduct a series of experiments to have a holistic comparison of six prevalent preference-based EMO algorithms against three iconic EMO algorithms without considering any preference information under various settings. In summary, we come up with the following five major observations. 
\begin{itemize}
    \item {\textit{A well designed EMO algorithm, without considering any DM's preference information, is able to be competitive for finding the SOI.}} This is particularly true when the number of objectives is small. However, this becomes significantly more difficult, if not impossible, with the increase of the number of objectives.
       % When the number of objectives is small, a well designed EMO algorithm, without considering any DM's preference information, is competitive for finding the SOI. However, this becomes significantly more difficult, if not impossible, with the increase of the number of objectives.
    \item {\textit{Dominance-, indicator- and decomposition-based frameworks all can be used as a baseline for designing effective preference-based EMO algorithms.}} Distance metric, such as Euclidean distance and Tchebycheff distance, is a reliable way to transform the DM supplied preference information into the selection pressure in an algorithm. On the other hand, if the DM's preference information is not well utilised, it bring more uncertainty to the search process thus make the end algorithm even worse than those non-preference-based counterparts.
    %Dominance-, indicator- and decomposition-based frameworks all can be used as a baseline for designing effective preference-based EMO algorithms, given that the DM supplied preference information is well utilised. Otherwise, using DM's preference information brings more uncertainty to the search process thus make the end algorithm even worse than those non-preference-based counterparts. In other words, using DM's preference information does not always guarantee a desirable approximation of the ROI.
    \item {\textit{A preference-based EMO algorithm may fail to find the ROI if the DM elicits an unreasonable preference information.}} Note that this is not uncommon when encountering a real-world black-box system of which the DM has little knowledge.
    \item {\textit{Interactive preference elicitation provide a better opportunity for the DM to progressively understand the underlying black-box system thus to gradually rectify her/his preference information.}}
    \item {\textit{Preference elicitation can be used as another means of decomposition method in EMO.}} That is to say, a preference-based EMO algorithm, e.g. R-NSGA-II, is able to approximate the whole PF given that the DM supplied preference information aim to cover the whole PF instead of a partial region.
\end{itemize}

EMO and MCDM are actually sibling communities which share many overlaps. However, they have been extensively developed in parallel in the past three decades. Although there is growing trend of seeking the synergy between them, as introduced in~\pref{sec:introduction}, it is still a lukewarm whilst more efforts are required along this line of research. This paper lights up the potential convergence between EMO and MCDM under the same paradigm. Many questions are still open for future exploration whilst we just name a few as follows:

\begin{itemize}
    \item This paper only investigate the case where the DM's preference information is represented as a reference point. It is interesting to investigate a universal framework that is able to embrace different preference information.

    \item Furthermore, it is far from trivial to quantitatively compare the performance of different preference-based EMO algorithms when they use different ways to represent preference information. In particular, the R-metric used in this paper can only be useful when a reference point is used to represent the DM's preference information. It is interesting to develop other performance metric for a wider range of preference representations.

    \item As discussed in~\pref{sec:interactive}, interactive EMO is a promising way to progressively approximate the SOI with the assistance of DM(s) under a black-box setting. It is worthwhile to invest more efforts to develop a human-in-the-loop optimisation paradigm in future.
\end{itemize}
%EMO and MCDM are two different communities but with many overlaps. In the past three decades, research in EMO mainly focus on finding a population of well converged and well diversified trade-off alternatives; whilst MCDM mainly aims to explicitly evaluate multiple conflicting criteria in decision-making and identify the most satisfactory solution with respect to the DM's preference information. As introduced in~\pref{sec:introduction}, there is a growing trend of seeking the synergy between EMO and MCDM. It is expected to see more work along this line of research in the near future. There are two major challenges ahead. The first one is the way used to represent the DM's preference information are versatile. This paper only investigate the case where the DM's preference information is represented as a reference point. It is interesting to investigate a universal framework that is able to embrace different kinds of preference elicitation methods. Followed by this, it is far from trivial to quantitatively compare the performance of different preference-based EMO algorithms when they use different way to represent preference information. In particular, the R-metric used in this paper can only be useful when a reference point is used to represent the DM's preference information. It is interesting to develop other performance metric for a wider range of preference representations. 

\bibliographystyle{IEEEtran}
\bibliography{IEEEabrv,casestudy}

\begin{thebibliography}{100}
\providecommand{\url}[1]{#1}
\csname url@samestyle\endcsname
\providecommand{\newblock}{\relax}
\providecommand{\bibinfo}[2]{#2}
\providecommand{\BIBentrySTDinterwordspacing}{\spaceskip=0pt\relax}
\providecommand{\BIBentryALTinterwordstretchfactor}{4}
\providecommand{\BIBentryALTinterwordspacing}{\spaceskip=\fontdimen2\font plus
\BIBentryALTinterwordstretchfactor\fontdimen3\font minus
  \fontdimen4\font\relax}
\providecommand{\BIBforeignlanguage}[2]{{%
\expandafter\ifx\csname l@#1\endcsname\relax
\typeout{** WARNING: IEEEtran.bst: No hyphenation pattern has been}%
\typeout{** loaded for the language `#1'. Using the pattern for}%
\typeout{** the default language instead.}%
\else
\language=\csname l@#1\endcsname
\fi
#2}}
\providecommand{\BIBdecl}{\relax}
\BIBdecl

\bibitem{DebAPM02}
K.~Deb, S.~Agrawal, A.~Pratap, and T.~Meyarivan, ``A fast and elitist
  multiobjective genetic algorithm: {NSGA-II},'' \emph{{IEEE} Trans.
  Evolutionary Computation}, vol.~6, no.~2, pp. 182--197, 2002.

\bibitem{ZitzlerK04}
E.~Zitzler and S.~K{\"{u}}nzli, ``Indicator-based selection in multiobjective
  search,'' in \emph{PPSN'04: Proc. of 8th International Conference on Parallel
  Problem Solving from Nature}, 2004, pp. 832--842.

\bibitem{ZhangL07}
Q.~Zhang and H.~Li, ``{MOEA/D:} {A} multiobjective evolutionary algorithm based
  on decomposition,'' \emph{{IEEE} Trans. Evolutionary Computation}, vol.~11,
  no.~6, pp. 712--731, 2007.

\bibitem{LiLTY15}
B.~Li, J.~Li, K.~Tang, and X.~Yao, ``Many-objective evolutionary algorithms:
  {A} survey,'' \emph{{ACM} Comput. Surv.}, vol.~48, no.~1, pp. 13:1--13:35,
  2015.

\bibitem{BrankeKS01}
J.~Branke, T.~Kaussler, and H.~Schmeck, ``Guidance in evolutionary
  multi-objective optimization,'' \emph{Advances in Engineering Software},
  vol.~32, no.~6, pp. 499--507, 2001.

\bibitem{Deb03}
K.~Deb, ``Multi-objective evolutionary algorithms: Introducing bias among
  pareto-optimal solutions,'' in \emph{Advances in Evolutionary Computing},
  A.~Ghosh and S.~Tsutsui, Eds.\hskip 1em plus 0.5em minus 0.4em\relax Springer
  Berlin Heidelberg, 2003, pp. 263--292.

\bibitem{BrankeD05}
J.~Branke and K.~Deb, ``Integrating user preferences into evolutionary
  multi-objective optimization,'' in \emph{Knowledge Incorporation in
  Evolutionary Computation}, Y.~Jin, Ed.\hskip 1em plus 0.5em minus 0.4em\relax
  Springer Berlin Heidelberg, 2005, vol. 167, pp. 461--477.

\bibitem{LuqueSHCC09}
J.~M. Luque, L.~V. Santana{-}Quintero, A.~G. Hern{\'{a}}ndez{-}D{\'{\i}}az,
  C.~A.~C. Coello, and R.~Caballero, ``g-dominance: Reference point based
  dominance for multiobjective metaheuristics,'' \emph{European Journal of
  Operational Research}, vol. 197, no.~2, pp. 685--692, 2009.

\bibitem{ThieleMKL09}
L.~Thiele, K.~Miettinen, P.~J. Korhonen, and J.~M. Luque, ``A preference-based
  evolutionary algorithm for multi-objective optimization,'' \emph{Evolutionary
  Computation}, vol.~17, no.~3, pp. 411--436, 2009.

\bibitem{DebK07GECCO}
K.~Deb and A.~Kumar, ``Interactive evolutionary multi-objective optimization
  and decision-making using reference direction method,'' in \emph{Genetic and
  Evolutionary Computation Conference, {GECCO} 2007, Proceedings, London,
  England, UK, July 7-11, 2007}, 2007, pp. 781--788.

\bibitem{DebK07CEC}
------, ``Light beam search based multi-objective optimization using
  evolutionary algorithms,'' in \emph{CEC'07: Proc. of the 2007 {IEEE} Congress
  on Evolutionary Computation}, 2007, pp. 2125--2132.

\bibitem{GreenwoodHD96}
G.~W. Greenwood, X.~Hu, and J.~G. D'Ambrosio, ``Fitness functions for multiple
  objective optimization problems: Combining preferences with pareto
  rankings,'' in \emph{FOGA'96: Proc. of the 4th Workshop on Foundations of
  Genetic Algorithms}, 1996, pp. 437--455.

\bibitem{PhelpsK03}
S.~P. Phelps and M.~K\"oksalan, ``An interactive evolutionary metaheuristic for
  multiobjective combinatorial optimization,'' \emph{Management Science},
  vol.~49, no.~12, pp. 1726--1738, 2003.

\bibitem{DebSKW10}
K.~Deb, A.~Sinha, P.~J. Korhonen, and J.~Wallenius, ``An interactive
  evolutionary multiobjective optimization method based on progressively
  approximated value functions,'' \emph{{IEEE} Trans. Evolutionary
  Computation}, vol.~14, no.~5, pp. 723--739, 2010.

\bibitem{LiDY17}
K.~Li, K.~Deb, and X.~Yao, ``R-metric: Evaluating the performance of
  preference-based evolutionary multiobjective optimization using reference
  points,'' \emph{{IEEE} Trans. Evolutionary Computation}, vol.~22, no.~6, pp.
  821--835, 2018.

\bibitem{Coello00}
C.~A.~C. Coello, ``Handling preference in evolution multiobjective
  optimization: A survey,'' in \emph{CEC'00: Proc. of the 2000 IEEE
  International Conference on Evolutionary Computation}, 2000, pp. 30--37.

\bibitem{RachmawatiS06}
L.~Rachmawati and D.~Srinivasan, ``Preference incorporation in multi-objective
  evolutionary algorithms: {A} survey,'' in \emph{CEC'06: Proc. of the 2006
  {IEEE} International Conference on Evolutionary Computation}, 2006, pp.
  962--968.

\bibitem{PurshouseDMMW14}
R.~C. Purshouse, K.~Deb, M.~M. Mansor, S.~Mostaghim, and R.~Wang, ``A review of
  hybrid evolutionary multiple criteria decision making methods,'' in
  \emph{CEC'14: Proc. of the 2014 {IEEE} Congress on Evolutionary Computation},
  2014, pp. 1147--1154.

\bibitem{BechikhKSG15}
S.~Bechikh, M.~Kessentini, L.~B. Said, and K.~Gh{\'{e}}dira, ``Preference
  incorporation in evolutionary multiobjective optimization: {A} survey of the
  state-of-the-art,'' in \emph{Advances in Computers}, 2015, vol.~98, pp.
  141--207.

\bibitem{SlowinskiGM02}
R.~Slowi\'nski and T.~Friedrich, ``Axiomatization of utility, outranking and
  decision rule preference models for multiple-criteria classification problems
  under partial inconsistency with the dominance principle,'' \emph{Control and
  Cybernetics}, vol.~31, pp. 1005--1035, 2002.

\bibitem{WickramasingheL08}
U.~K. Wickramasinghe and X.~Li, ``Integrating user preferences with particle
  swarms for multi-objective optimization,'' in \emph{GECCO'08: Proc. of the
  2008 Genetic and Evolutionary Computation Conference}, 2008, pp. 745--752.

\bibitem{GongLZJZ11}
M.~Gong, F.~Liu, W.~Zhang, L.~Jiao, and Q.~Zhang, ``Interactive {MOEA/D} for
  multi-objective decision making,'' in \emph{GECCO'11: Proc. of the 13th
  Annual Genetic and Evolutionary Computation Conference}, 2011, pp. 721--728.

\bibitem{LiCMY18}
K.~Li, R.~Chen, G.~Min, and X.~Yao, ``Integration of preferences in
  decomposition multiobjective optimization,'' \emph{{IEEE} Trans.
  Cybernetics}, vol.~48, no.~12, pp. 3359--3370, 2018.

\bibitem{LiCSY18}
K.~Li, R.~Chen, D.~A. Savic, and X.~Yao, ``Interactive decomposition
  multiobjective optimization via progressively learned value functions,''
  \emph{{IEEE} Trans. Fuzzy Systems}, vol.~27, no.~5, pp. 849--860, 2019.

\bibitem{FonsecaF98}
C.~M. Fonseca and P.~J. Fleming, ``Multiobjective optimization and multiple
  constraint handling with evolutionary algorithms. i. {A} unified
  formulation,'' \emph{{IEEE} Trans. Systems, Man, and Cybernetics, Part {A}},
  vol.~28, no.~1, pp. 26--37, 1998.

\bibitem{DebSBC06}
K.~Deb, J.~Sundar, U.~Bhaskara, and S.~Chaudhuri, ``Reference point based
  multiobjective optimization using evolutionary algorithms,''
  \emph{International Journal of Computational Intelligence Research}, vol.~2,
  no.~3, pp. 273--286, 2006.

\bibitem{DebK07}
K.~Deb and A.~Kumar, ``Interactive evolutionary multi-objective optimization
  and decision-making using reference direction method,'' in \emph{Genetic and
  Evolutionary Computation Conference, {GECCO} 2007, Proceedings, London,
  England, UK, July 7-11, 2007}, 2007, pp. 781--788.

\bibitem{AllmendingerLB08}
R.~Allmendinger, X.~Li, and J.~Branke, ``Reference point-based particle swarm
  optimization using a steady-state approach,'' in \emph{SEAL'08: Proc. of the
  7th International Conference on Simulated Evolution and Learning}, 2008, pp.
  200--209.

\bibitem{MolinaSDCC09}
J.~Molina, L.~V. Santana, A.~G. Hern\'andez-D\'iaz, C.~A.~C. Coello, and
  R.~Caballero, ``g-dominance: Reference point based dominance for
  multiobjective metaheuristics,'' \emph{European Journal of Operational
  Research}, vol. 197, pp. 685--692, 2009.

\bibitem{SaidBG10}
L.~B. Said, S.~Bechikh, and K.~Gh{\'{e}}dira, ``The r-dominance: {A} new
  dominance relation for interactive evolutionary multicriteria decision
  making,'' \emph{{IEEE} Trans. Evolutionary Computation}, vol.~14, no.~5, pp.
  801--818, 2010.

\bibitem{MohammadiOLD14}
A.~Mohammadi, M.~N. Omidvar, X.~Li, and K.~Deb, ``Integrating user preferences
  and decomposition methods for many-objective optimization,'' in \emph{CEC'14:
  Proc. of the 2014 {IEEE} Congress on Evolutionary Computation}, 2014, pp.
  421--428.

\bibitem{Yu2015}
G.~Yu, J.~Zheng, R.~Shen, and M.~Li, ``Decomposing the user-preference in
  multiobjective optimization,'' \emph{Soft Computing}, pp. 1--17, 2015.

\bibitem{Ma2015}
X.~Ma, F.~Liu, Y.~Qi, L.~Li, L.~Jiao, X.~D.~X. Wang, B.~Dong, Z.~Hou, Y.~Zhang,
  and J.~Wu, ``{MOEA/D} with biased weight adjustment inspired by user
  preference and its application on multi-objective reservoir flood control
  problem,'' \emph{Soft Computing}, pp. 1--25, 2015.

\bibitem{NarukawaSTOSI16}
K.~Narukawa, Y.~Setoguchi, Y.~Tanigaki, M.~Olhofer, B.~Sendhoff, and
  H.~Ishibuchi, ``Preference representation using gaussian functions on a
  hyperplane in evolutionary multi-objective optimization,'' \emph{Soft
  Comput.}, vol.~20, no.~7, pp. 2733--2757, 2016.

\bibitem{TrautmannWB13}
H.~Trautmann, T.~Wagner, and D.~Brockhoff, ``{R2-EMOA:} focused multiobjective
  search using {R}2-indicator-based selection,'' in \emph{LION 7: Proc. of the
  7th International Conference on Learning and Intelligent Optimization}.\hskip
  1em plus 0.5em minus 0.4em\relax Springer, 2013, pp. 70--74.

\bibitem{WagnerTB13}
T.~Wagner, H.~Trautmann, and D.~Brockhoff, ``Preference articulation by means
  of the {R2} indicator,'' in \emph{EMO'13: Proc. of the 7th International
  Conference on Evolutionary Multi-Criterion Optimization}.\hskip 1em plus
  0.5em minus 0.4em\relax Springer, 2013, pp. 81--95.

\bibitem{ZitzlerBT06}
E.~Zitzler, D.~Brockhoff, and L.~Thiele, ``The {Hypervolume} indicator
  revisited: On the design of {Pareto}-compliant indicators via weighted
  integration,'' in \emph{EMO'07: Proc. of the 4th International Conference on
  Evolutionary Multi-Criterion Optimization}, 2006, pp. 862--876.

\bibitem{FriedrichKN13}
T.~Friedrich, T.~Kroeger, and F.~Neumann, ``Weighted preferences in
  evolutionary multi-objective optimization,'' \emph{Int. J. Machine Learning
  {\&} Cybernetics}, vol.~4, no.~2, pp. 139--148, 2013.

\bibitem{ZitzlerLT01}
E.~Zitzler, M.~Laumanns, and L.~Thiele, ``{SPEA2}: Improving the strength
  pareto evolutionary algorithm,'' Computer Engineering and Networks Laboratory
  (TIK), ETH, Zurich, Switzerland, Tech. Rep. TIK Report 103, 2001.

\bibitem{WagnerT10}
T.~Wagner and H.~Trautmann, ``Integration of preferences in hypervolume-based
  multiobjective evolutionary algorithms by means of desirability functions,''
  \emph{{IEEE} Trans. Evolutionary Computation}, vol.~14, no.~5, pp. 688--701,
  2010.

\bibitem{BeumeNE07}
N.~Beume, B.~Naujoks, and M.~T.~M. Emmerich, ``{SMS-EMOA:} multiobjective
  selection based on dominated hypervolume,'' \emph{European Journal of
  Operational Research}, vol. 181, no.~3, pp. 1653--1669, 2007.

\bibitem{ParmeeC02}
D.~Cvetkovi\'c and I.~C. Parmee, ``Preferences and their application in
  evolutionary multiobjective optimization,'' \emph{{IEEE} Trans. Evolutionary
  Computation}, vol.~6, no.~1, pp. 42--57, 2002.

\bibitem{JinS02}
Y.~Jin and B.~Sendhoff, ``Incorporation of fuzzy preferences into evolutionary
  multiobjective optimization,'' in \emph{{GECCO'02}: Proc. of the 2002 Genetic
  and Evolutionary Computation Conference}, 2002, p. 683.

\bibitem{ShenGCH10}
X.~Shen, Y.~Guo, Q.~Chen, and W.~Hu, ``A multi-objective optimization
  evolutionary algorithm incorporating preference information based on fuzzy
  logic,'' \emph{Comp. Opt. and Appl.}, vol.~46, no.~1, pp. 159--188, 2010.

\bibitem{BattitiP10}
R.~Battiti and A.~Passerini, ``Brain-computer evolutionary multiobjective
  optimization: {A} genetic algorithm adapting to the decision maker,''
  \emph{{IEEE} Trans. Evolutionary Computation}, vol.~14, no.~5, pp. 671--687,
  2010.

\bibitem{BrankeGSZ15}
J.~Branke, S.~Greco, R.~Slowi\'nski, and P.~Zielniewicz, ``Learning value
  functions in interactive evolutionary multiobjective optimization,''
  \emph{{IEEE} Trans. Evolutionary Computation}, vol.~19, no.~1, pp. 88--102,
  2015.

\bibitem{PedroT14}
L.~R. Pedro and R.~H.~C. Takahashi, ``{INSPM:} an interactive evolutionary
  multi-objective algorithm with preference model,'' \emph{Inf. Sci.}, vol.
  268, pp. 202--219, 2014.

\bibitem{GrecoMS11}
S.~Greco, B.~Matarazzo, and R.~Slowi\'nski, ``Interactive multiobjective
  mixed-integer optimization using dominance-based rough set approach,'' in
  \emph{EMO'11: Proc. of the 6th International Conference Evolutionary
  Multi-Criterion Optimization}, 2011, pp. 241--253.

\bibitem{MiettinenM00}
K.~Miettinen and M.~M. M{\"{a}}kel{\"{a}}, ``Interactive multiobjective
  optimization system {WWW-NIMBUS} on the internet,'' \emph{Computers {\&}
  {OR}}, vol.~27, no. 7-8, pp. 709--723, 2000.

\bibitem{MiettinenERL10}
K.~Miettinen, P.~Eskelinen, F.~Ruiz, and M.~Luque, ``{NAUTILUS} method: An
  interactive technique in multiobjective optimization based on the nadir
  point,'' \emph{European Journal of Operational Research}, vol. 206, no.~2,
  pp. 426--434, 2010.

\bibitem{SindhyaRM11}
K.~Sindhya, A.~B. Ruiz, and K.~Miettinen, ``A preference based interactive
  evolutionary algorithm for multi-objective optimization: {PIE},'' in
  \emph{EMO'11: Proc. of the 6th International Conference Evolutionary
  Multi-Criterion Optimization}, 2011, pp. 212--225.

\bibitem{Yang99}
J.~Yang, ``Gradient projection and local region search for multiobjective
  optimisation,'' \emph{European Journal of Operational Research}, vol. 112,
  no.~2, pp. 432--459, 1999.

\bibitem{YangL02}
J.~Yang and D.~Li, ``Normal vector identification and interactive tradeoff
  analysis using minimax formulation in multiobjective optimization,''
  \emph{{IEEE} Trans. Systems, Man, and Cybernetics, Part {A}}, vol.~32, no.~3,
  pp. 305--319, 2002.

\bibitem{LuqueYW09}
M.~Luque, J.~Yang, and B.~Y.~H. Wong, ``{PROJECT} method for multiobjective
  optimization based on gradient projection and reference points,''
  \emph{{IEEE} Trans. Systems, Man, and Cybernetics, Part {A}}, vol.~39, no.~4,
  pp. 864--879, 2009.

\bibitem{ChenXC17}
L.~Chen, B.~Xin, and J.~Chen, ``A trade-off based interactive multi-objective
  optimization method driven by evolutionary algorithms,'' \emph{J. Adv.
  Comput. Intell. Intell. Inform.}, vol.~21, no.~2, pp. 284--292, 2017.

\bibitem{FowlerGKKMW10}
J.~W. Fowler, E.~S. Gel, M.~K{\"{o}}ksalan, P.~J. Korhonen, J.~L. Marquis, and
  J.~Wallenius, ``Interactive evolutionary multi-objective optimization for
  quasi-concave preference functions,'' \emph{European Journal of Operational
  Research}, vol. 206, no.~2, pp. 417--425, 2010.

\bibitem{SinhaKWD10}
A.~Sinha, P.~J. Korhonen, J.~Wallenius, and K.~Deb, ``An interactive
  evolutionary multi-objective optimization method based on polyhedral cones,''
  in \emph{LION IV: Proc. of the 4th International Conference Learning and
  Intelligent Optimization}, 2010, pp. 318--332.

\bibitem{KoksalanK10}
M.~K{\"{o}}ksalan and I.~Karahan, ``An interactive territory defining
  evolutionary algorithm: itdea,'' \emph{{IEEE} Trans. Evolutionary
  Computation}, vol.~14, no.~5, pp. 702--722, 2010.

\bibitem{KarahanK10}
I.~Karahan and M.~K{\"{o}}ksalan, ``A territory defining multiobjective
  evolutionary algorithms and preference incorporation,'' \emph{{IEEE} Trans.
  Evolutionary Computation}, vol.~14, no.~4, pp. 636--664, 2010.

\bibitem{DebG11}
K.~Deb and S.~Gupta, ``Understanding knee points in bicriteria problems and
  their implications as preferred solution principles,'' \emph{Engineering
  Optimization}, vol.~43, no.~11, pp. 1175--1204, 2011.

\bibitem{BhattacharjeeSR17}
K.~S. Bhattacharjee, H.~K. Singh, M.~Ryan, and T.~Ray, ``Bridging the gap:
  Many-objective optimization and informed decision-making,'' \emph{{IEEE}
  Trans. Evolutionary Computation}, vol.~21, no.~5, pp. 813--820, 2017.

\bibitem{EversonWF14}
R.~M. Everson, D.~J. Walker, and J.~E. Fieldsend, ``Life on the edge:
  Characterising the edges of mutually non-dominating sets,''
  \emph{Evolutionary Computation}, vol.~22, no.~3, pp. 479--501, 2014.

\bibitem{SinghBR18}
H.~K. Singh, K.~S. Bhattacharjee, and T.~Ray, ``Distance based subset selection
  for benchmarking in evolutionary multi/many-objective optimization,''
  \emph{{IEEE} Trans. Evolutionary Computation}, 2018, accepted for
  publication.

\bibitem{IshibuchiISN17}
H.~Ishibuchi, R.~Imada, Y.~Setoguchi, and Y.~Nojima, ``Hypervolume subset
  selection for triangular and inverted triangular pareto fronts of
  three-objective problems,'' in \emph{FOGA'17: Proc. of the 14th {ACM/SIGEVO}
  Conference on Foundations of Genetic Algorithms}, 2017, pp. 95--110.

\bibitem{BosmanT03}
P.~A.~N. Bosman and D.~Thierens, ``The balance between proximity and diversity
  in multiobjective evolutionary algorithms,'' \emph{{IEEE} Trans. Evolutionary
  Computation}, vol.~7, no.~2, pp. 174--188, 2003.

\bibitem{ZitzlerT99}
E.~Zitzler and L.~Thiele, ``Multiobjective evolutionary algorithms: a
  comparative case study and the strength pareto approach,'' \emph{{IEEE}
  Trans. Evolutionary Computation}, vol.~3, no.~4, pp. 257--271, 1999.

\bibitem{LiZZL09}
K.~Li, J.~Zheng, C.~Zhou, and H.~Lv, ``An improved differential evolution for
  multi-objective optimization,'' in \emph{CSIE'09: Proc. of 2009 {WRI} World
  Congress on Computer Science and Information Engineering}, 2009, pp.
  825--830.

\bibitem{LiZLZL09}
K.~Li, J.~Zheng, M.~Li, C.~Zhou, and H.~Lv, ``A novel algorithm for
  non-dominated hypervolume-based multiobjective optimization,'' in
  \emph{SMC'09: Proc. of 2009 the {IEEE} International Conference on Systems,
  Man and Cybernetics}, 2009, pp. 5220--5226.

\bibitem{LiFK11}
K.~Li, {\'{A}}.~Fialho, and S.~Kwong, ``Multi-objective differential evolution
  with adaptive control of parameters and operators,'' in \emph{LION5: Proc. of
  the 5th International Conference on Learning and Intelligent Optimization},
  2011, pp. 473--487.

\bibitem{LiKWCR12}
K.~Li, S.~Kwong, R.~Wang, J.~Cao, and I.~J. Rudas, ``Multi-objective
  differential evolution with self-navigation,'' in \emph{SMC'12: Proc. of the
  2012 {IEEE} International Conference on Systems, Man, and Cybernetics}, 2012,
  pp. 508--513.

\bibitem{CaoKWL12}
J.~Cao, S.~Kwong, R.~Wang, and K.~Li, ``A weighted voting method using minimum
  square error based on extreme learning machine,'' in \emph{ICMLC'12: Proc. of
  the 2012 International Conference on Machine Learning and Cybernetics}, 2012,
  pp. 411--414.

\bibitem{LiKWTM13}
K.~Li, S.~Kwong, R.~Wang, K.~Tang, and K.~Man, ``Learning paradigm based on
  jumping genes: {A} general framework for enhancing exploration in
  evolutionary multiobjective optimization,'' \emph{Inf. Sci.}, vol. 226, pp.
  1--22, 2013.

\bibitem{LiWKC13}
K.~Li, R.~Wang, S.~Kwong, and J.~Cao, ``Evolving extreme learning machine
  paradigm with adaptive operator selection and parameter control,''
  \emph{International Journal of Uncertainty, Fuzziness and Knowledge-Based
  Systems}, vol.~21, pp. 143--154, 2013.

\bibitem{LiK14}
K.~Li and S.~Kwong, ``A general framework for evolutionary multiobjective
  optimization via manifold learning,'' \emph{Neurocomputing}, vol. 146, pp.
  65--74, 2014.

\bibitem{CaoKWL14}
J.~Cao, S.~Kwong, R.~Wang, and K.~Li, ``{AN} indicator-based selection
  multi-objective evolutionary algorithm with preference for multi-class
  ensemble,'' in \emph{ICMLC'14: Proc. of the 2014 International Conference on
  Machine Learning and Cybernetics}, 2014, pp. 147--152.

\bibitem{LiZKLW14}
K.~Li, Q.~Zhang, S.~Kwong, M.~Li, and R.~Wang, ``Stable matching-based
  selection in evolutionary multiobjective optimization,'' \emph{{IEEE} Trans.
  Evolutionary Computation}, vol.~18, no.~6, pp. 909--923, 2014.

\bibitem{LiFKZ14}
K.~Li, {\'{A}}.~Fialho, S.~Kwong, and Q.~Zhang, ``Adaptive operator selection
  with bandits for a multiobjective evolutionary algorithm based on
  decomposition,'' \emph{{IEEE} Trans. Evolutionary Computation}, vol.~18,
  no.~1, pp. 114--130, 2014.

\bibitem{CaoKWLLK15}
J.~Cao, S.~Kwong, R.~Wang, X.~Li, K.~Li, and X.~Kong, ``Class-specific soft
  voting based multiple extreme learning machines ensemble,''
  \emph{Neurocomputing}, vol. 149, pp. 275--284, 2015.

\bibitem{WuKZLWL15}
M.~Wu, S.~Kwong, Q.~Zhang, K.~Li, R.~Wang, and B.~Liu, ``Two-level stable
  matching-based selection in {MOEA/D},'' in \emph{SMC'15: Proc. of the 2015
  {IEEE} International Conference on Systems, Man, and Cybernetics}, 2015, pp.
  1720--1725.

\bibitem{LiDZK15}
K.~Li, K.~Deb, Q.~Zhang, and S.~Kwong, ``An evolutionary many-objective
  optimization algorithm based on dominance and decomposition,'' \emph{{IEEE}
  Trans. Evolutionary Computation}, vol.~19, no.~5, pp. 694--716, 2015.

\bibitem{LiKZD15}
K.~Li, S.~Kwong, Q.~Zhang, and K.~Deb, ``Interrelationship-based selection for
  decomposition multiobjective optimization,'' \emph{{IEEE} Trans.
  Cybernetics}, vol.~45, no.~10, pp. 2076--2088, 2015.

\bibitem{LiKD15}
K.~Li, S.~Kwong, and K.~Deb, ``A dual-population paradigm for evolutionary
  multiobjective optimization,'' \emph{Inf. Sci.}, vol. 309, pp. 50--72, 2015.

\bibitem{LiDZZ17}
K.~Li, K.~Deb, Q.~Zhang, and Q.~Zhang, ``Efficient nondomination level update
  method for steady-state evolutionary multiobjective optimization,''
  \emph{{IEEE} Trans. Cybernetics}, vol.~47, no.~9, pp. 2838--2849, 2017.

\bibitem{WuKJLZ17}
M.~Wu, S.~Kwong, Y.~Jia, K.~Li, and Q.~Zhang, ``Adaptive weights generation for
  decomposition-based multi-objective optimization using gaussian process
  regression,'' in \emph{Proceedings of the Genetic and Evolutionary
  Computation Conference, {GECCO} 2017, Berlin, Germany, July 15-19, 2017},
  2017, pp. 641--648.

\bibitem{WuLKZZ17}
M.~Wu, K.~Li, S.~Kwong, Y.~Zhou, and Q.~Zhang, ``Matching-based selection with
  incomplete lists for decomposition multiobjective optimization,''
  \emph{{IEEE} Trans. Evolutionary Computation}, vol.~21, no.~4, pp. 554--568,
  2017.

\bibitem{LiDY18}
K.~Li, K.~Deb, and X.~Yao, ``R-metric: Evaluating the performance of
  preference-based evolutionary multiobjective optimization using reference
  points,'' \emph{{IEEE} Trans. Evolutionary Computation}, vol.~22, no.~6, pp.
  821--835, 2018.

\bibitem{ChenLY18}
R.~Chen, K.~Li, and X.~Yao, ``Dynamic multiobjectives optimization with a
  changing number of objectives,'' \emph{{IEEE} Trans. Evolutionary
  Computation}, vol.~22, no.~1, pp. 157--171, 2018.

\bibitem{WuLKZ18}
M.~Wu, K.~Li, S.~Kwong, and Q.~Zhang, ``Evolutionary many-objective
  optimization based on adversarial decomposition,'' \emph{{IEEE} Trans.
  Cybernetics}, 2018, accepted for publication.

\bibitem{ChenLBY18}
T.~Chen, K.~Li, R.~Bahsoon, and X.~Yao, ``{FEMOSAA:} feature-guided and
  knee-driven multi-objective optimization for self-adaptive software,''
  \emph{{ACM} Trans. Softw. Eng. Methodol.}, vol.~27, no.~2, pp. 5:1--5:50,
  2018.

\bibitem{LiCFY19}
K.~Li, R.~Chen, G.~Fu, and X.~Yao, ``Two-archive evolutionary algorithm for
  constrained multiobjective optimization,'' \emph{{IEEE} Trans. Evolutionary
  Computation}, vol.~23, no.~2, pp. 303--315, 2019.

\bibitem{WuLKZZ19}
M.~Wu, K.~Li, S.~Kwong, Q.~Zhang, and J.~Zhang, ``Learning to decompose: {A}
  paradigm for decomposition-based multiobjective optimization,'' \emph{{IEEE}
  Trans. Evolutionary Computation}, vol.~23, no.~3, pp. 376--390, 2019.

\bibitem{LiCSY19}
K.~Li, R.~Chen, D.~A. Savic, and X.~Yao, ``Interactive decomposition
  multiobjective optimization via progressively learned value functions,''
  \emph{{IEEE} Trans. Fuzzy Systems}, vol.~27, no.~5, pp. 849--860, 2019.

\bibitem{Li19}
K.~Li, ``Progressive preference learning: Proof-of-principle results in
  {MOEA/D},'' in \emph{EMO'19: Proc. of the 10th International Conference
  Evolutionary Multi-Criterion Optimization}, 2019, pp. 631--643.

\bibitem{BillingsleyLMMG19}
J.~Billingsley, K.~Li, W.~Miao, G.~Min, and N.~Georgalas, ``A formal model for
  multi-objective optimisation of network function virtualisation placement,''
  in \emph{EMO'19: Proc. of the 10th International Conference Evolutionary
  Multi-Criterion Optimization}, 2019, pp. 529--540.

\bibitem{ZouJYZZL19}
J.~Zou, C.~Ji, S.~Yang, Y.~Zhang, J.~Zheng, and K.~Li, ``A knee-point-based
  evolutionary algorithm using weighted subpopulation for many-objective
  optimization,'' \emph{Swarm and Evolutionary Computation}, vol.~47, pp.
  33--43, 2019.

\bibitem{DebJ14}
K.~Deb and H.~Jain, ``An evolutionary many-objective optimization algorithm
  using reference-point-based nondominated sorting approach, part {I:} solving
  problems with box constraints,'' \emph{{IEEE} Trans. Evolutionary
  Computation}, vol.~18, no.~4, pp. 577--601, 2014.

\bibitem{ZitzlerDT00}
E.~Zitzler, K.~Deb, and L.~Thiele, ``Comparison of multiobjective evolutionary
  algorithms: Empirical results,'' \emph{Evolutionary Computation}, vol.~8,
  no.~2, pp. 173--195, 2000.

\bibitem{DebTLZ05}
K.~Deb, L.~Thiele, M.~Laumanns, and E.~Zitzler, \emph{Scalable Test Problems
  for Evolutionary Multiobjective Optimization}.\hskip 1em plus 0.5em minus
  0.4em\relax London: Springer London, 2005, pp. 105--145.

\bibitem{IshibuchiAN15}
H.~Ishibuchi, N.~Akedo, and Y.~Nojima, ``Behavior of multiobjective
  evolutionary algorithms on many-objective knapsack problems,'' \emph{{IEEE}
  Trans. Evolutionary Computation}, vol.~19, no.~2, pp. 264--283, 2015.

\bibitem{GiagkiozisPF13}
I.~Giagkiozis, R.~C. Purshouse, and P.~J. Fleming, ``Towards understanding the
  cost of adaptation in decomposition-based optimization algorithms,'' in
  \emph{SMC'13: Proc. of the 2013 {IEEE} International Conference on Systems,
  Man, and Cybernetics}, 2013, pp. 615--620.

\bibitem{KonnoS95}
H.~Konno and K.~ichi Suzuki, ``A mean-variance-skewness portfolio optimization
  model,'' \emph{Journal of the Operations Research Society of Japan}, vol.~38,
  no.~2, pp. 173--187, 1995.

\bibitem{LaiYW06}
K.~K. Lai, L.~Yu, and S.~Wang, ``Mean-variance-skewness-kurtosis-based
  portfolio optimization,'' in \emph{IMSCCS'06: Proc. of the 1st International
  Multi-Symposium of Computer and Computational Sciences}, 2006, pp. 292--297.

\bibitem{LiDAY17}
K.~Li, K.~Deb, O.~T. Altin{\"{o}}z, and X.~Yao, ``Empirical investigations of
  reference point based methods when facing a massively large number of
  objectives: First results,'' in \emph{EMO'17: Proc. of the 9th International
  Conference Evolutionary Multi-Criterion Optimization}, 2017, pp. 390--405.

\bibitem{BringmannF13}
K.~Bringmann and T.~Friedrich, ``Approximation quality of the hypervolume
  indicator,'' \emph{Artif. Intell.}, vol. 195, pp. 265--290, 2013.

\end{thebibliography}

\end{document}